\documentclass[11pt,a4paper]{article}

\usepackage[margin=25mm]{geometry}
\usepackage[utf8]{inputenc}
\usepackage[T1]{fontenc}
\usepackage{lmodern}
\usepackage{microtype}
\usepackage{graphicx}
\graphicspath{{figures/}}
\usepackage{amsmath,amssymb}
\usepackage{booktabs}
\usepackage{makecell}
\usepackage{rotating}
\usepackage{array}
\usepackage{float}
\usepackage[section]{placeins}
\usepackage{caption}
\usepackage{authblk}
\usepackage[numbers,sort&compress,super]{natbib}   
\usepackage{lineno}
\usepackage[table]{xcolor}
\usepackage{xurl}
\usepackage[colorlinks=true,linkcolor=blue!60!black,citecolor=blue!60!black,urlcolor=blue!60!black]{hyperref}

\newcommand{\co}{CO$_2$}
\newcommand{\etco}{EtCO$_2$}
\newcommand{\spo}{SpO$_2$}
\newcommand{\dmodel}{d_{\mathrm{model}}}

\title{\bfseries A Cardiorespiratory Foundation Model Pretrained on Ten Physiological Signals for Perioperative and Critical Care}

\author[1]{Cheol-Hui Lee}
\author[2]{Wookyung Kim}
\author[3]{Hyeonhoon Lee}
\author[1]{Garam Lee}
\author[2]{Da-in Eun}
\author[1,4,*]{Hyung-Chul~Lee}

\affil[1]{Healthcare AI Research Institute, Seoul National University Hospital, Seoul, Republic of Korea}
\affil[2]{Interdisciplinary Program in Medical Informatics, College of Medicine, Seoul National University, Seoul, Republic of Korea}
\affil[3]{Department of Transdisciplinary Medicine, Seoul National University Hospital, Seoul, Republic of Korea}
\affil[4]{Department of Anesthesiology and Pain Medicine, Seoul National University College of Medicine and Seoul National University Hospital, Seoul, Republic of Korea}
\affil[*]{Corresponding author: vital@snu.ac.kr}

\date{}

\begin{document}
\maketitle
{\renewcommand{\thefootnote}{}\footnotetext{\url{https://github.com/dlcjfgmlnasa/CARMEN}}}

\begin{abstract}
\noindent Arterial, venous, pulmonary artery, and intracranial pressures and ventilator waveforms guide decisions in operating rooms and intensive care units, yet foundation models for physiological waveforms rely almost entirely on the electrocardiogram and photoplethysmogram. We present the CArdio-Respiratory Multi-modal ENcoder (CARMEN), a single encoder pretrained without labels on ten cardiorespiratory waveforms (2.27 million modality-hours). CARMEN accepts any combination of co-recorded signals at variable lengths. A conditioning mechanism, LSCNorm, restores the absolute signal level removed by normalization; in a matched ablation, replacing it with standard normalization lowered the AUROC for intraoperative hypotension by 0.11. Across 19 tasks from Korean and US cohorts, the frozen encoder significantly outperformed both supervised models in 12 of the 15 tasks on which they were evaluated. It outperformed three pretrained foundation models on 14--17 of 19 tasks, and none of them exceeded it on any task. Given the electrocardiogram or photoplethysmogram alone, it matched or exceeded the signal-specific foundation models in most comparisons, and the performance was largely preserved even after removing the secondary channels.
\end{abstract}


\section{Introduction}\label{sec:intro}

Intensive care units (ICUs) and operating rooms (ORs) are the clinical settings where physiological signals are most densely recorded, and the interpretation of these signals most directly influences patient outcomes. Invasive hemodynamic waveforms, such as arterial, central venous, and intracranial pressures, are recorded continuously together with respiratory waveforms. These signals are associated with a wide range of clinical outcomes, from acute events unfolding within minutes to organ failure and death occurring over days or weeks \cite{VitalDB, MIMICIIIWDB, Salmasi2017}. Delayed or inappropriate interventions increase morbidity, mortality, and long-term functional decline \cite{Needham2012}. However, limited clinical resources and the cognitive load on clinicians make it difficult to interpret the trajectories and interactions of multiple signals at the bedside \cite{Goldart2024}. Deep learning models developed to support this task have mostly remained supervised designs specialized for individual signals and tasks. They require large-scale labeling for every new task \cite{Rajpurkar2022} and fail to learn a unified representation that spans multiple tasks and modalities \cite{Krishnan2022}.

Foundation models (FMs) built on self-supervised pretraining have overcome this limitation. They learn a single representation from large unlabeled corpora and transfer it to many tasks. FMs are replacing task-specific supervised models across medical domains, including medical imaging, pathology, and electronic health records \cite{RETFound, UNI, NYUTron}, and have also been proposed for biosignals \cite{HeartBEiT, PaPaGei, BIOT}. Recently, the cardiac sensing foundation model (CSFM) learned 12-lead and single-lead electrocardiogram (ECG) and photoplethysmogram (PPG) signals within one encoder and showed that a single model can handle many cardiac tasks even when the combination of modalities changes \cite{CSFM}. FMs pretrained on ICU recordings have also emerged, but at the time of writing, remain confined to ECG, PPG, and peripheral oxygen saturation (\spo) \cite{QualityFM, MorphologyFM}. Biosignal FMs now accommodate non-invasive cardiac signals, but not the invasive hemodynamic and respiratory waveforms of the ICU and OR, where the benefit of an FM would be greatest. Three limitations create this gap.

First, pretraining modalities have been restricted to non-invasive signals with abundant public data such as ECG and PPG. Invasive hemodynamic signals have been effectively excluded from pretraining; these include arterial blood pressure (ABP), central venous pressure (CVP), pulmonary artery pressure (PAP), and intracranial pressure (ICP). So have respiratory signals, including capnography (\co), airway pressure (AWP), respiratory flow (RespFlow), and impedance respiration (RespImp). In other words, the direction of model development has been dictated by data availability rather than clinical need. Second, most existing FMs accept only a single short, fixed-length signal, or a predetermined combination of modalities. This makes them difficult to apply in real monitoring environments, where sensor changes and signal dropout are routine. Third, signal normalization removes the absolute scale of the signal. Normalization is widely used to reduce variation across patients and measurement devices. For example, consider two arterial pressure waveforms of identical shape with mean arterial pressures (MAPs) of 55 mmHg and 85 mmHg, respectively. After normalization, they may become indistinguishable. The absolute blood pressure information on which clinicians base their decisions to intervene is thus not conveyed to the model.

To address these three limitations, we propose the CArdio-Respiratory Multi-modal ENcoder (CARMEN), a multimodal FM specialized for continuous cardiorespiratory signals in the ICU and OR. First, CARMEN learns six cardiovascular signals (ECG, ABP, PPG, CVP, PAP, and ICP) and four respiratory signals (\co, AWP, RespImp, and RespFlow) in a unified representation space using a single encoder. To our knowledge, this is the first biosignal FM that incorporates both invasive hemodynamic and respiratory waveforms into pretraining. Second, CARMEN accepts any combination of available signals and variable-length inputs without imputing missing channels. Therefore, a single encoder can be applied in clinical environments where the monitoring configurations differ from patient to patient (Fig.~\ref{fig:overview}a). Third, we introduce Loc/Scale-conditioned Layer Normalization (LSCNorm), which re-injects the mean and standard deviation removed by normalization as conditioning inputs for each layer of the encoder. Therefore, the absolute levels of the signals with physical units are retained in the representation.

CARMEN was pretrained on three datasets. Two are bedside monitor waveform recordings collected over 21 months from all ORs and four ICUs in a tertiary hospital. The third is K-MIMIC-MORTAL, a registry of deceased ICU patients from the same hospital. The model was evaluated on two cohorts: a Korean OR cohort that was not used for pretraining and a US cohort to assess generalization across institutions and countries. The evaluation covered 19 downstream tasks in five categories: clinical event detection, short-term event prediction, admission outcome prediction, physiological estimation, and phenotype.

\section{Results}\label{sec:results}

\subsection{Study population and downstream tasks}\label{sec:res-population}

CARMEN was pretrained on three cohorts from Seoul National University Hospital (SNUH): two continuous-waveform cohorts from all ORs and four ICUs, plus K-MIMIC-MORTAL. The inputs were ten waveforms: six cardiovascular (ECG, PPG, ABP, CVP, PAP, and ICP) and four respiratory (\co, AWP, RespImp, and RespFlow). The pretraining data comprised 526,061 records (331,913 from the SNUH OR and ICU cohorts and 194,148 from 725 patients in K-MIMIC-MORTAL). Summed across modalities, this amounts to 2,274,701 modality-hours, approximately 260 years of signal and 33 billion tokens (Fig.~\ref{fig:overview}b). Invasive hemodynamic and respiratory waveforms accounted for 21.7\% and 28.2\% of these modality-hours, respectively. These are waveforms recorded by bedside monitors during routine care and are not records selected for research. Therefore, they span a wide range of physiological deterioration, from intraoperative hemodynamic fluctuations to the ICU course of patients who died. Three model sizes (Small, 52M; Base, 174M; Large, 410M) were pretrained on this corpus. Unless otherwise stated, the results are reported for the Base model.

There were 19 downstream tasks. Nine were derived from the OR cohort of the same institution (SNUH) as the pretraining data, and ten were derived from the US cohorts to assess generalization across institutions and countries. All evaluation cohorts were excluded from the pretraining data (Fig.~\ref{fig:overview}c). The inputs ranged from a single signal to combinations of up to four signals, and the window lengths ranged from 15 s to 10 min, depending on the task. Short-term event prediction forecasts physiological events minutes after the observation window, whereas admission outcome prediction estimates the clinical outcome of the entire ICU stay from the first hours of recording (Fig.~\ref{fig:overview}c). Seven of the 19 tasks relied on signals that the existing biosignal FMs did not handle. Four took invasive hemodynamic waveforms as their primary input (Intraoperative Hypotension, Massive Transfusion, Intracranial Hypertension, and Cardiac Output) and three took respiratory waveforms (Hypoxemia, Sleep Apnea, and End-tidal \co{} (\etco) Derangement).

\begin{figure}[p]
\centering
\makebox[\textwidth][c]{%
\includegraphics[height=\dimexpr\paperheight-30mm\relax,width=\dimexpr\paperwidth-40mm\relax,keepaspectratio]{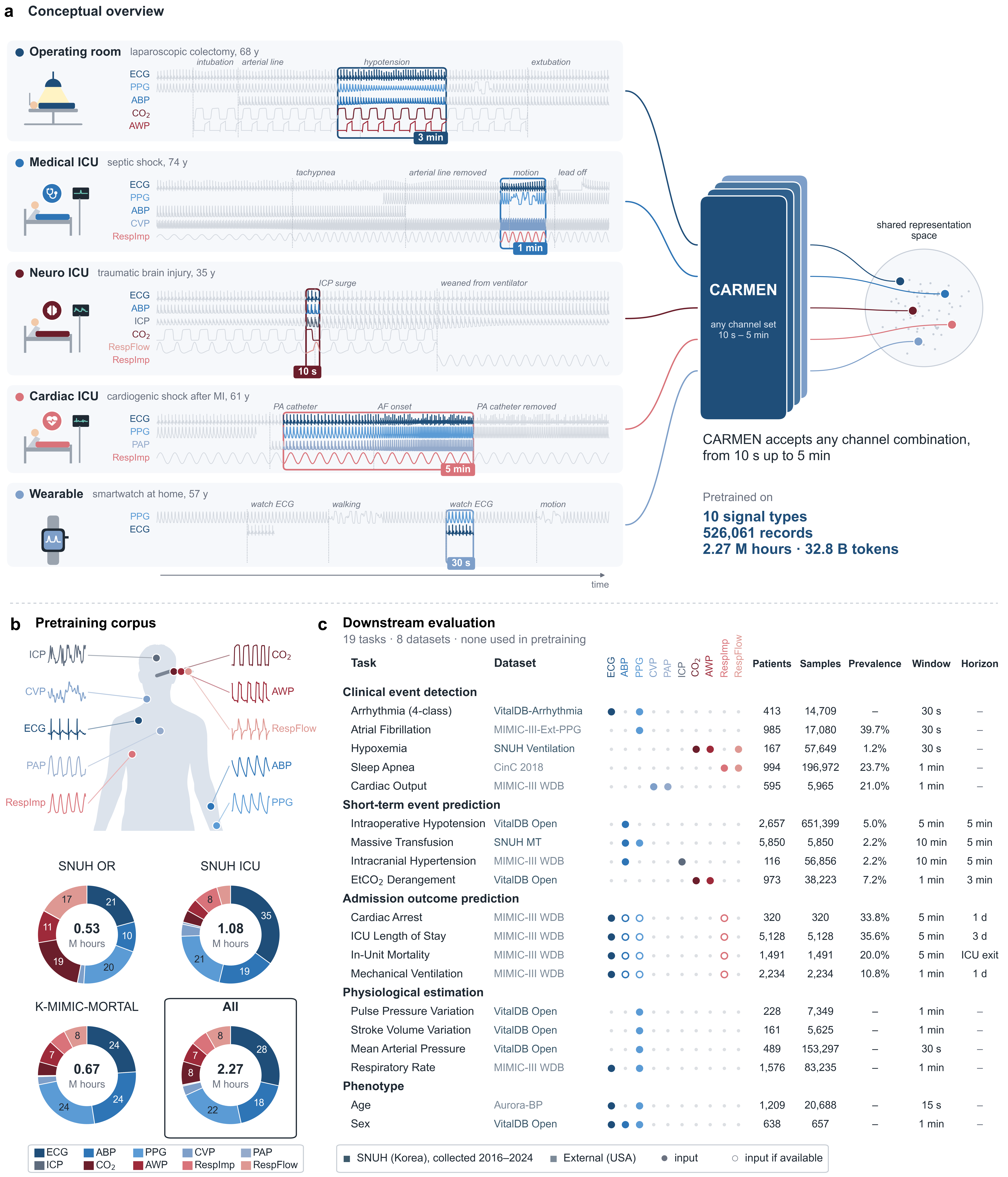}}
\caption{\textbf{Overview of CARMEN.} Legend on the following page.}
\label{fig:overview}
\end{figure}

\begin{figure}[!t]
\ContinuedFloat
\caption{\textbf{Overview of CARMEN} (continued). \textbf{a}, Concept. The recorded signals differ between care settings and patients and change over time. CARMEN encodes whichever signals are recorded together in a window (boxes; 10 s to 5 min in pretraining, up to 10 min in downstream tasks) with a single encoder, without imputation. \textbf{b}, Pretraining corpus. Top, measurement sites and example waveforms; bottom, share of recording time per signal in each cohort (center, millions of hours). \textbf{c}, The 19 downstream tasks. Prevalence, positive rate (--, multi-class, continuous-target and Sex tasks); Horizon, time from the end of observation to the event (to ICU exit for In-Unit Mortality). ECG, electrocardiogram; PPG, photoplethysmogram; ABP, arterial blood pressure; CVP, central venous pressure; PAP, pulmonary artery pressure; ICP, intracranial pressure; \co, capnogram; AWP, airway pressure; RespImp, impedance respiration; RespFlow, respiratory flow; \etco, end-tidal \co; SNUH, Seoul National University Hospital; OR, operating room; ICU, intensive care unit.}\end{figure}

\subsection{Downstream task performance}\label{sec:res-performance}

CARMEN, with a frozen encoder and a trained linear layer (linear probe), was compared with five comparators over 19 tasks (Fig.~\ref{fig:performance}a, d). Two were supervised: CARMEN-scratch uses the same encoder architecture trained end-to-end from random initialization, providing a comparison between frozen pretrained representations and direct supervised training of the same architecture; 1D-ResNet is a widely used supervised baseline for biosignals \cite{He2016ResNet}. Three were pretrained: MOMENT \cite{MOMENT}, a general-purpose time-series FM, and NormWear \cite{NormWear} and SPOTR \cite{SPOTR}, biosignal FMs, all evaluated with the same linear probe as CARMEN. NormWear and SPOTR accept only short fixed-length inputs; therefore, each task window was split into consecutive segments of the length used in their pretraining, and the segment embeddings were averaged to give one embedding per window. The four admission outcome prediction tasks, which had small samples, were compared only with pretrained models, without supervised comparators. Classification was evaluated using the area under the receiver operating characteristic curve (AUROC) and regression using the mean absolute error (MAE). Significance was defined as $q<0.05$ after a paired patient-cluster bootstrap with Benjamini--Hochberg correction (see Methods).

With only one linear layer trained, CARMEN outperformed the supervised models trained end-to-end for most tasks (Fig.~\ref{fig:performance}d). For 12 of the 15 tasks with supervised comparators, CARMEN significantly outperformed both. There were three exceptions: for Hypoxemia and Intracranial Hypertension, CARMEN did not differ significantly from CARMEN-scratch. For MAP estimation, 1D-ResNet had a significantly lower error; however, the difference was only 0.30 mmHg.

CARMEN also outperformed the pretrained FMs evaluated using the same linear probe (Fig.~\ref{fig:performance}d). Of the 19 tasks, CARMEN was significantly superior to MOMENT on 16, NormWear on 14, and SPOTR on 17, and none of the three significantly outperformed CARMEN on any task. The largest gaps were observed for Intracranial Hypertension, where CARMEN exceeded MOMENT, NormWear, and SPOTR by 0.265, 0.313, and 0.319 in AUROC, and for Intraoperative Hypotension, where it exceeded them by 0.231, 0.234, and 0.244, respectively. Both tasks were defined by the absolute thresholds of invasive pressures. On both, CARMEN-scratch, which shares CARMEN's architecture, came within 0.01 of CARMEN (0.865 vs.\ 0.870 and 0.888 vs.\ 0.895). This indicates that these gaps stem mainly from the architecture's use of invasive pressure waveforms and their absolute levels rather than from pretraining. Comparisons without a significant difference were concentrated in the admission outcome prediction tasks, where small samples yielded wide confidence intervals; the remainder were for Hypoxemia, Cardiac Output, and MAP.

\begin{figure}[p]
\centering
\makebox[\textwidth][c]{%
\includegraphics[height=\dimexpr\paperheight-30mm\relax,width=\dimexpr\paperwidth-40mm\relax,keepaspectratio]{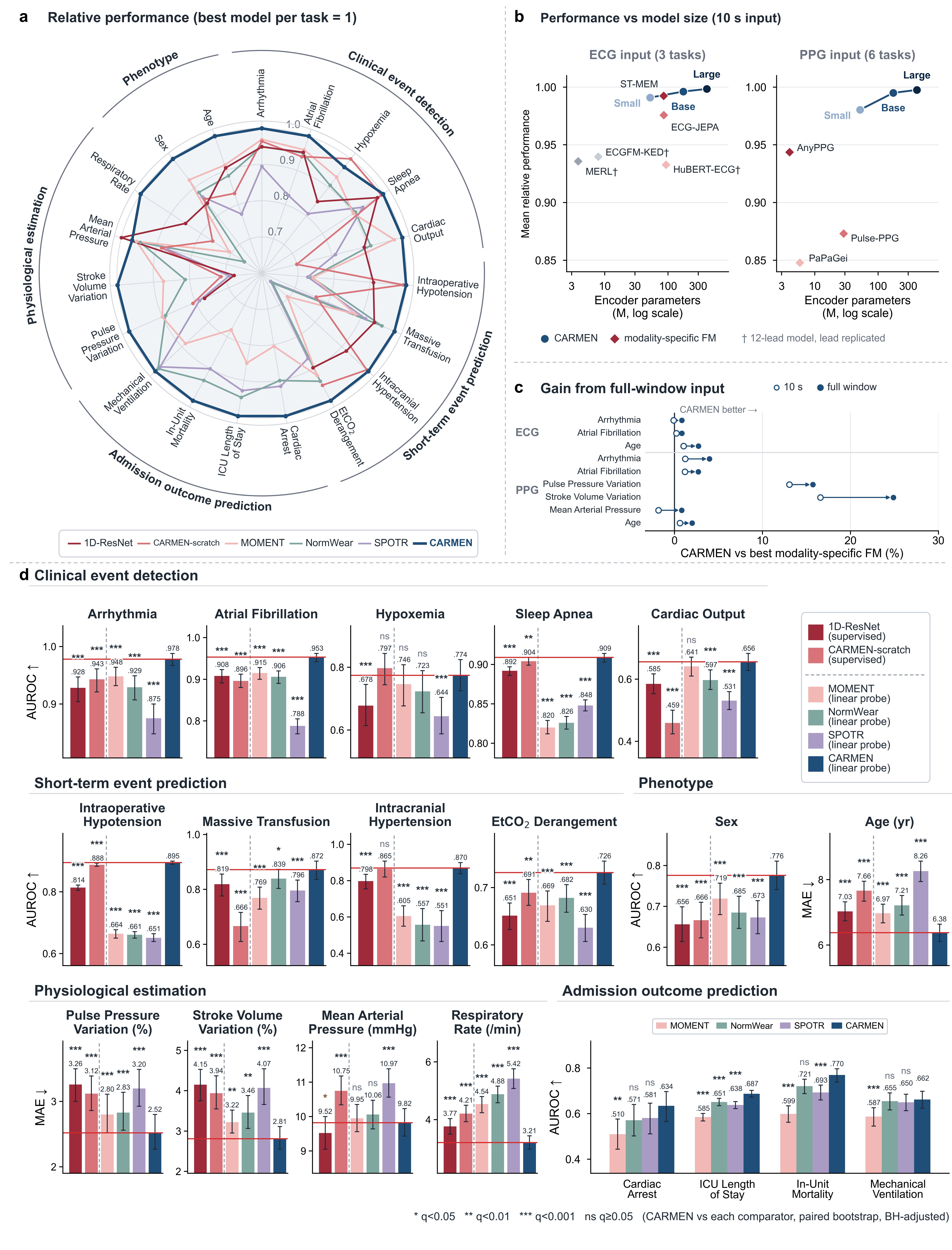}}
\caption{\textbf{Downstream task performance (CARMEN-Base, 174M).} Legend on the following page.}
\label{fig:performance}
\end{figure}

\begin{figure}[!t]
\ContinuedFloat
\caption{\textbf{Downstream task performance (CARMEN-Base, 174M)} (continued). \textbf{a}, Performance on the 19 tasks relative to the best model on each task (AUROC, or 1/MAE for regression). Supervised models were not trained for admission outcome prediction. \textbf{b}, Mean relative performance against encoder size for CARMEN at three sizes (10 s input) and the signal-specific FMs. \textbf{c}, Gain of CARMEN over the best signal-specific FM on each task, with 10 s input and with the full window. \textbf{d}, Performance on each task with 95\% CIs; the red line marks CARMEN. Significance is against CARMEN (paired patient-cluster bootstrap, Benjamini--Hochberg adjusted), and brown symbols mark a comparator that was better. AUROC, area under the receiver operating characteristic curve; MAE, mean absolute error; CI, confidence interval; FM, foundation model.}
\end{figure}

\begin{table}[!tb]
\caption{\textbf{Comparison with existing biosignal FMs.} All models were evaluated with the same patient-level 5-fold linear probe. In each input block, the upper rows compare CARMEN-Base at 10 s input with each FM at its own pretraining input length (cropped from the window center), and the shaded row gives CARMEN with the full window. Values are AUROC (macro-averaged for Arrhythmia) or MAE (PPV, SVV, MAP and Age) with patient-cluster bootstrap 95\% CIs; bold, best among the upper rows; underlined, second best. ECG and ECG+PPG inputs for Atrial Fibrillation were added for this comparison. $\dagger$, 12-lead models given the single available lead replicated across the 12 leads. PPV, pulse pressure variation; SVV, stroke volume variation; MAP, mean arterial pressure; AUROC, area under the receiver operating characteristic curve; MAE, mean absolute error; CI, confidence interval.}\label{tab:fm}
\small
\setlength{\tabcolsep}{5pt}
\textbf{(a) ECG-based input}\par\medskip
\resizebox{\textwidth}{!}{\begin{tabular}{llcccc}
\toprule
Model & Modality & Input & \makecell{Arrhythmia\\(Macro-AUROC)} & \makecell{Atrial Fibrillation\\(AUROC)} & \makecell{Age\\(MAE, yr)} \\
\midrule
CARMEN & ECG+PPG & 10 s & \textbf{0.967} (0.953--0.979) & \textbf{0.964} (0.956--0.972) & \textbf{6.45} (6.18--6.73) \\
NormWear \cite{NormWear} & ECG+PPG & 6 s & \underline{0.914} (0.894--0.934) & \underline{0.924} (0.911--0.936) & \underline{7.21} (6.92--7.52) \\
SPOTR \cite{SPOTR} & ECG+PPG & 10 s & 0.865 (0.842--0.886) & 0.854 (0.838--0.870) & 8.40 (8.07--8.71) \\
BIOT \cite{BIOT} & ECG+PPG & 10 s & 0.828 (0.794--0.861) & 0.772 (0.749--0.796) & 7.90 (7.58--8.20) \\
\rowcolor{gray!15} CARMEN & ECG+PPG & Full window & 0.978 (0.967--0.988) & 0.968 (0.960--0.976) & 6.38 (6.11--6.64) \\
\midrule
CARMEN & ECG & 10 s & \underline{0.967} (0.957--0.977) & \textbf{0.963} (0.955--0.971) & \textbf{7.67} (7.37--7.98) \\
ST-MEM \cite{STMEM} & ECG & 9 s & \textbf{0.968} (0.959--0.977) & \underline{0.961} (0.951--0.970) & \underline{7.75} (7.44--8.06) \\
ECG-JEPA \cite{ECGJEPA} & ECG & 10 s & 0.964 (0.951--0.975) & 0.960 (0.949--0.970) & 8.12 (7.79--8.45) \\
HuBERT-ECG$\dagger$ \cite{HuBERTECG} & ECG & 10 s & 0.928 (0.909--0.946) & 0.915 (0.901--0.929) & 8.53 (8.20--8.87) \\
ECGFM-KED$\dagger$ \cite{ECGFMKED} & ECG & 10 s & 0.931 (0.909--0.951) & 0.934 (0.920--0.946) & 8.55 (8.21--8.89) \\
MERL$\dagger$ \cite{MERL} & ECG & 10 s & 0.899 (0.872--0.923) & 0.934 (0.919--0.947) & 8.35 (8.02--8.68) \\
\rowcolor{gray!15} CARMEN & ECG & Full window & 0.976 (0.965--0.985) & 0.969 (0.961--0.976) & 7.54 (7.25--7.84) \\
\bottomrule
\end{tabular}}
\medskip

\textbf{(b) PPG input}\par\medskip
{\footnotesize\setlength{\tabcolsep}{4pt}
\resizebox{\textwidth}{!}{\begin{tabular}{lccccccc}
\toprule
Model & Input & \makecell{Arrhythmia\\(Macro-AUROC)} & \makecell{Atrial Fibrillation\\(AUROC)} & \makecell{PPV\\(MAE, \%)} & \makecell{SVV\\(MAE, \%)} & \makecell{MAP\\(MAE, mmHg)} & \makecell{Age\\(MAE, yr)} \\
\midrule
CARMEN & 10 s & \makecell{\textbf{0.917}\\(0.884--0.944)} & \makecell{\textbf{0.939}\\(0.928--0.950)} & \makecell{\textbf{2.60}\\(2.38--2.84)} & \makecell{\textbf{3.12}\\(2.88--3.39)} & \makecell{\underline{10.08}\\(9.68--10.48)} & \makecell{\textbf{6.49}\\(6.22--6.76)} \\
AnyPPG \cite{AnyPPG} & 10 s & \makecell{\underline{0.906}\\(0.864--0.939)} & \makecell{\underline{0.928}\\(0.915--0.941)} & \makecell{\underline{2.99}\\(2.74--3.28)} & \makecell{\underline{3.74}\\(3.37--4.12)} & \makecell{\textbf{9.90}\\(9.51--10.30)} & \makecell{\underline{6.53}\\(6.26--6.80)} \\
PaPaGei \cite{PaPaGei} & 10 s & \makecell{0.818\\(0.770--0.861)} & \makecell{0.839\\(0.823--0.854)} & \makecell{3.07\\(2.82--3.35)} & \makecell{4.10\\(3.71--4.49)} & \makecell{11.06\\(10.66--11.46)} & \makecell{8.01\\(7.68--8.30)} \\
Pulse-PPG \cite{PulsePPG} & Full window & \makecell{0.820\\(0.782--0.853)} & \makecell{0.864\\(0.847--0.878)} & \makecell{3.04\\(2.76--3.36)} & \makecell{3.88\\(3.49--4.29)} & \makecell{10.53\\(10.12--10.95)} & \makecell{7.76\\(7.45--8.06)} \\
\rowcolor{gray!15} CARMEN & Full window & \raisebox{-1.1ex}[2.9ex][2.0ex]{\shortstack[c]{0.942\\(0.912--0.964)}} & \raisebox{-1.1ex}[2.9ex][2.0ex]{\shortstack[c]{0.953\\(0.943--0.963)}} & \raisebox{-1.1ex}[2.9ex][2.0ex]{\shortstack[c]{2.52\\(2.27--2.80)}} & \raisebox{-1.1ex}[2.9ex][2.0ex]{\shortstack[c]{2.81\\(2.55--3.11)}} & \raisebox{-1.1ex}[2.9ex][2.0ex]{\shortstack[c]{9.82\\(9.43--10.23)}} & \raisebox{-1.1ex}[2.9ex][2.0ex]{\shortstack[c]{6.40\\(6.13--6.66)}} \\
\bottomrule
\end{tabular}}}
\end{table}

\subsection{Comparison with existing biosignal foundation models}\label{sec:res-fm}

CARMEN was then compared with existing biosignal FMs on ECG and PPG---the modalities covered by public FMs---across six tasks evaluable from these two signals: Arrhythmia, Atrial Fibrillation, pulse pressure variation (PPV), stroke volume variation (SVV), MAP, and Age (Fig.~\ref{fig:performance}b and Table~\ref{tab:fm}). Eleven FMs were included: five ECG-only models (ST-MEM \cite{STMEM}, ECG-JEPA \cite{ECGJEPA}, HuBERT-ECG \cite{HuBERTECG}, ECGFM-KED \cite{ECGFMKED}, and MERL \cite{MERL}); three PPG-only models (AnyPPG \cite{AnyPPG}, PaPaGei \cite{PaPaGei}, and Pulse-PPG \cite{PulsePPG}); and three models that take ECG and PPG together (BIOT \cite{BIOT}, NormWear \cite{NormWear}, and SPOTR \cite{SPOTR}). Each comparator received inputs of the length used in its own pretraining, whereas CARMEN received ECG, PPG, and ECG+PPG inputs from a single set of pretrained weights. CARMEN was first evaluated on 10 s inputs alongside the comparators whose input lengths are listed in Table~\ref{tab:fm}; results with the full original window are reported separately.

With 10 s inputs, CARMEN was the best in 10 of the 12 combinations of input modality and task (Table~\ref{tab:fm}). Of the other two, Arrhythmia with ECG input did not differ significantly from ST-MEM ($\Delta$AUROC $-$0.001, $q=0.85$), and on MAP with PPG input, AnyPPG's MAE was lower by 0.18 mmHg (less than 2\% of the MAE, $q=0.014$). The gap was largest for PPV and SVV, which reflect respiratory beat-to-beat variation; CARMEN's MAE was 13.0\% and 16.6\% lower than AnyPPG's (2.60 vs.\ 2.99\%; 3.12 vs.\ 3.74\%). With ECG+PPG input, CARMEN surpassed NormWear, the strongest of the three multi-signal FMs, on all three tasks (Arrhythmia 0.967 vs.\ 0.914; Atrial Fibrillation 0.964 vs.\ 0.924; Age MAE 6.45 vs.\ 7.21 yr).

With the full original window, CARMEN outperformed its 10-s-input version in all 12 combinations and was the best for all six tasks (Fig.~\ref{fig:performance}c and Table~\ref{tab:fm}). The two combinations that had trailed at 10 s also surpassed the dedicated FMs, although on MAP the difference from AnyPPG was not significant (9.82 vs.\ 9.90 mmHg, $q=0.59$). The MAE reduction relative to AnyPPG increased to 15.7\% for PPV and 24.9\% for SVV. The comparators were evaluated at their own pretraining lengths and not on the full window. This comparison shows what CARMEN gains from longer context and is not a like-for-like ranking at equal input lengths.

\subsection{Robustness to missing channels}\label{sec:res-realworld}

\begin{figure}[!tb]
\centering
\includegraphics[width=0.9\textwidth]{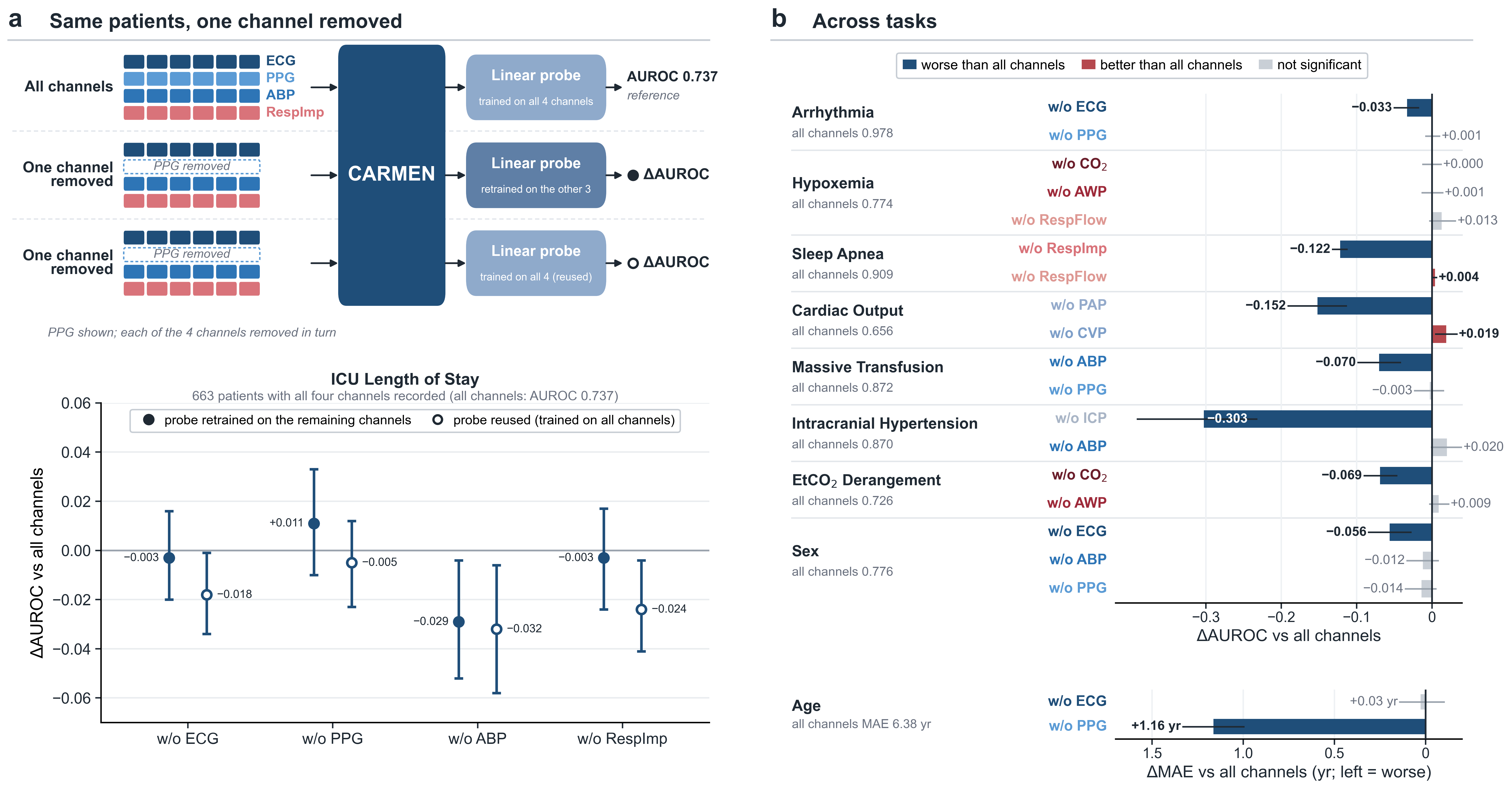}
\caption{\textbf{Robustness to missing channels.} \textbf{a}, ICU Length of Stay in the 663 patients with all four channels recorded. Top, the three conditions; bottom, change in AUROC when each channel is removed in turn. \textbf{b}, The same analysis in the other multi-channel tasks, with the probe retrained (for Age, change in MAE; left is worse). Error bars, 95\% CIs (paired patient-cluster bootstrap). AUROC, area under the receiver operating characteristic curve; MAE, mean absolute error; CI, confidence interval.}\label{fig:realworld}\end{figure}

Unlike curated benchmarks, real clinical data differ in which signals are recorded for each patient: across the four admission outcome prediction tasks, only 6.4--16.6\% of patients had all of ECG, PPG, ABP, and RespImp recorded (Fig.~\ref{fig:overview}c). CARMEN encodes only the recorded channels as a single sequence; therefore, each patient's actual channels were input without imputing missing channels or excluding patients. The effect of missing channels was assessed by removing one channel at a time in 663 ICU Length of Stay patients with all four channels. Removing the ECG, PPG, or RespImp left AUROC unchanged (0.733--0.748 vs.\ 0.737 for all channels; $\Delta$ $-$0.003 to +0.011), and removing ABP alone lowered it to 0.708 ($\Delta$ $-$0.029, 95\% CI $-$0.052 to $-$0.004; Fig.~\ref{fig:realworld}a). Even when the probe trained on all channels was retained and a channel was removed only at evaluation, the decrease was limited to 0.005--0.032.

We extended the same analysis to other tasks with multi-channel input (Fig.~\ref{fig:realworld}b). Performance fell markedly only when the physiological signal that defined the task was removed, and removing the other channels rarely changed it. AUROC decreased significantly upon removing ECG in Arrhythmia ($\Delta$ $-$0.033), ABP in Massive Transfusion ($-$0.070), PAP in Cardiac Output ($-$0.152), ICP in Intracranial Hypertension ($-$0.303), \co{} in \etco{} Derangement ($-$0.069), RespImp in Sleep Apnea ($-$0.122), and ECG in Sex ($-$0.056), whereas removing the remaining channels of the same tasks did not. In Hypoxemia, removing any one of \co, AWP, or RespFlow made no significant difference ($\Delta$ +0.000 to +0.013), suggesting that the three respiratory signals carry redundant information. Removing CVP in Cardiac Output (+0.019) and RespFlow in Sleep Apnea (+0.004) increased AUROC significantly, if slightly. For Age, removing PPG increased the MAE by 1.16 years, whereas removing ECG made no significant difference (+0.03 years).

\subsection{Ablation studies}\label{sec:res-ablation}

\begin{figure}[!tb]
\centering
\includegraphics[width=\textwidth]{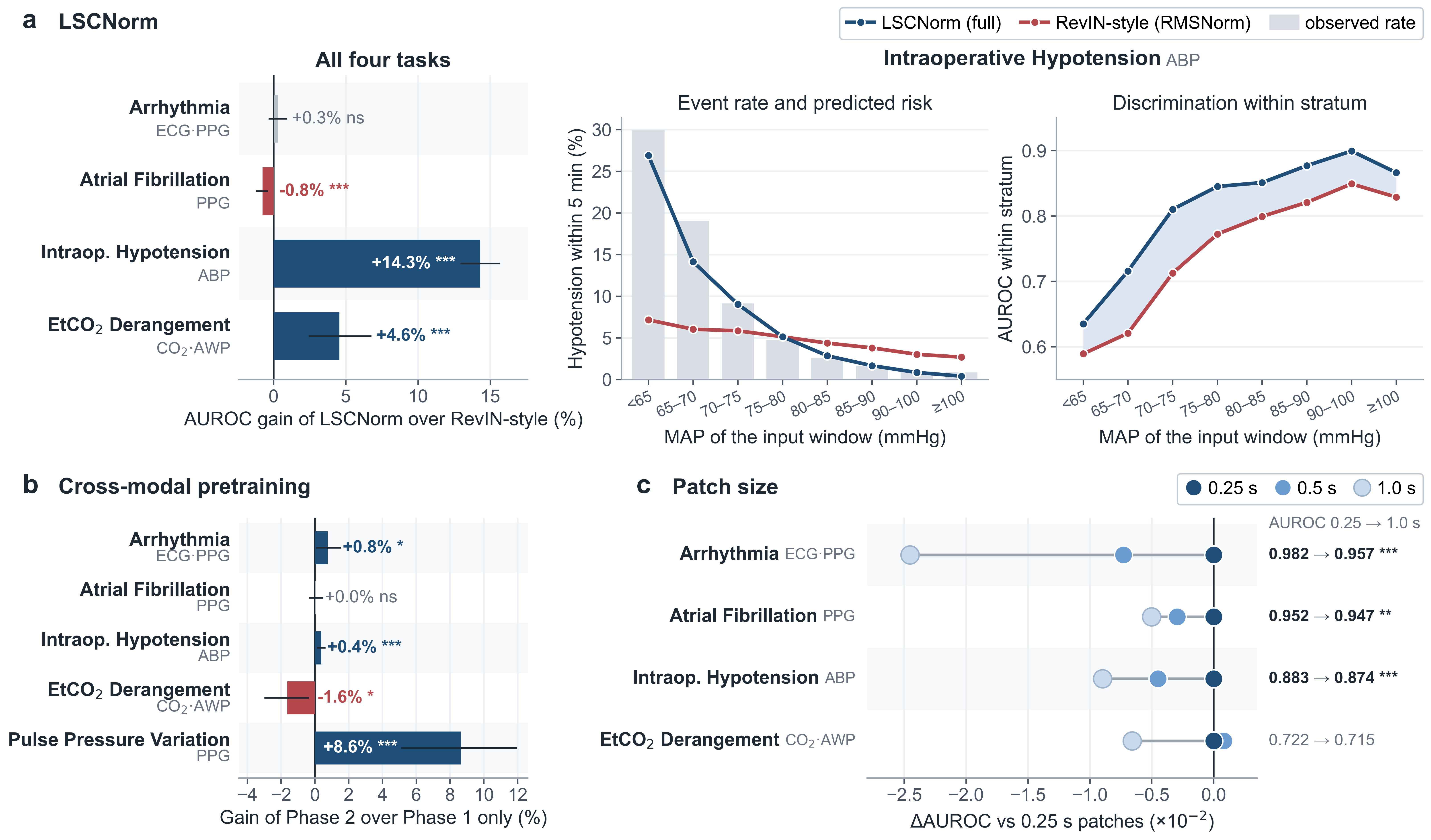}
\caption{\textbf{Ablation studies.} Each variant changes one design element of a reference model trained with the same budget: a model trained with about half the tokens of CARMEN-Base in \textbf{a} and \textbf{b}, and a separate model in a reduced setting in \textbf{c} (Methods). \textbf{a}, LSCNorm replaced by RMSNorm (RevIN-style: the removed mean and standard deviation are not returned to the encoder). Left, relative gain in AUROC from LSCNorm. Middle and right, Intraoperative Hypotension stratified by the MAP of the input window: observed event rate (bars) with mean predicted risk (lines), and AUROC within each stratum. These two panels use cross-validated predictions made for the ablation. \textbf{b}, Relative gain of the model trained through Phase 2 over the Phase 1 checkpoint (MAE for Pulse Pressure Variation, AUROC otherwise). \textbf{c}, Change in AUROC with 0.5 s and 1 s patches relative to 0.25 s patches, in the reduced setting. Horizontal lines, 95\% CIs; navy, reference model better; red, variant better; gray, not significant (paired patient-cluster bootstrap, Benjamini--Hochberg adjusted; * $q<0.05$, ** $q<0.01$, *** $q<0.001$). LSCNorm, Loc/Scale-conditioned Layer Normalization; RMSNorm, root mean square normalization; RevIN, reversible instance normalization; MAP, mean arterial pressure; AUROC, area under the receiver operating characteristic curve; MAE, mean absolute error; CI, confidence interval.}\label{fig:ablation}
\end{figure}

The contributions of CARMEN's three design elements---LSCNorm, cross-modal pretraining, and patch size---were evaluated (Fig.~\ref{fig:ablation}). LSCNorm and patch size were evaluated for four tasks chosen based on whether the absolute level of the signal determines the label. Arrhythmia and Atrial Fibrillation are defined by the waveform rhythm and are independent of the absolute level. Intraoperative Hypotension and \etco{} Derangement are defined by absolute thresholds on arterial pressure and end-tidal \co, respectively. Cross-modal pretraining was evaluated on five tasks, these four plus PPV, which requires reading beat-to-beat variations across the respiratory cycle. Because changing LSCNorm or the patch size requires pretraining from scratch, the ablations were run at smaller budgets than CARMEN-Base, and each variant was compared with a reference model trained with the same data and budget. The reference model for LSCNorm and Phase 2 was trained with about half the tokens of CARMEN-Base, and patch size was compared in a reduced setting with its own reference (about 15\% of that budget; Methods). The reference values therefore differ from those in Supplementary Tables \hyperref[tab:s-detection]{S2}--\hyperref[tab:s-phenotype]{S6}; those of the reference model for LSCNorm and Phase 2 nonetheless lay within the 95\% CIs of CARMEN-Base on Intraoperative Hypotension, \etco{} Derangement, Atrial Fibrillation, and PPV. The conclusions rest on the paired difference between each variant and its reference.

The gain from LSCNorm was confined to two tasks defined by absolute thresholds (Fig.~\ref{fig:ablation}a). Compared with the model in which LSCNorm was replaced by root mean square normalization (RMSNorm), the AUROC of the reference model was 14.3\% higher for Intraoperative Hypotension (0.894 vs.\ 0.782) and 4.6\% higher for \etco{} Derangement (0.736 vs.\ 0.704). There was no significant difference in Arrhythmia (+0.3\%). For Atrial Fibrillation, the RMSNorm model was slightly but significantly higher ($-$0.8\%; 0.950 vs.\ 0.957). The conditioning path was switched off for PPG, the only input of this task (see Methods); therefore, this difference does not arise from the injected statistics.

To determine the origin of this difference, we stratified the Intraoperative Hypotension input windows using MAP. The predicted risk of the LSCNorm model tracked the observed hypotension rate across the MAP strata. In contrast, the RMSNorm model varied little between strata. Consequently, it greatly underestimated the risk of windows in which the pressure was already low (below 65 mmHg: observed 29.9\%, LSCNorm 26.9\%, RMSNorm 7.2\%). The within-stratum AUROC was also higher for the LSCNorm model in every stratum.

Completing Phase 2 changed performance the most in the task that required reading beat-to-beat variations (Fig.~\ref{fig:ablation}b). Relative to the Phase 1 checkpoint, it reduced the PPV MAE by 8.6\% (2.42\% vs.\ 2.65\%). AUROC was also significantly higher on Arrhythmia (+0.8\%) and Intraoperative Hypotension (+0.4\%). On Atrial Fibrillation, there was no difference (+0.0\%). For \etco{} Derangement, the reference model's AUROC was 1.6\% lower.

Larger patches reduced performance most on Arrhythmia (Fig.~\ref{fig:ablation}c). In the reduced setting, increasing the patch from 0.25 s to 1 s significantly lowered AUROC on Arrhythmia (0.982 to 0.957), Atrial Fibrillation (0.952 to 0.947), and Intraoperative Hypotension (0.883 to 0.874). On \etco{} Derangement, there was no significant difference (0.722 to 0.715). On the first three tasks, the 0.5 s patch fell between the two.

\section{Discussion}\label{sec:discussion}

The waveforms recorded most densely in the OR and ICU have been the least extensively modeled. Invasive hemodynamic and respiratory waveforms are written to bedside monitors continuously, yet they have been left out of deep learning for lack of labels, because models have traditionally been developed for one signal, one event, and one labeled dataset at a time. CARMEN inverts this order. It first learns a single representation, without labels, from the ten cardiorespiratory waveforms that hospitals record every day; each task is then added on top with a single linear layer. With its encoder frozen, CARMEN outperformed task-specific supervised models and three pretrained FMs on most of the 19 downstream tasks, none of which were included in pretraining (Fig.~\ref{fig:performance}a, d; Supplementary Tables \hyperref[tab:s-detection]{S2}--\hyperref[tab:s-phenotype]{S6}).

The size of the gains is less notable than the reach of a single representation. The 19 tasks differ in input signals, observation length, and prediction horizon (Fig.~\ref{fig:overview}c), yet all used the same pretrained weights, and only a linear layer was trained per task. Minute-scale acute events, day-scale prognosis, physiological regression, and patient attributes were all read from one encoder. The representation also generalized across institutions. Pretrained at a single Korean hospital, it showed gains over comparators on most of the ten tasks from US cohorts (Fig.~\ref{fig:performance}d; Supplementary Table \hyperref[tab:datasets]{S1}). The encoder was reused without additional pretraining; only the task-specific linear probes were trained on the US data.

The representation also spans a longer time scale than existing biosignal FMs. Ten of the eleven FMs we compared were pretrained on fixed segments of 10 s or less, and their input is limited to that length (Table~\ref{tab:fm}). CARMEN was pretrained on continuous records with segment lengths varying from 10 s to 5 min, and in downstream tasks it processed the window defined by each task, up to 10 min, without modification. Position was encoded relatively, with no fixed-length position table; therefore, the encoder itself imposed no limit on window length. A corpus of 2.27 million modality-hours, approximately 260 years of signal laid end to end, was large enough to expose the model to such long windows in sufficient variety. The effect was visible when only the input length on the same task was changed. The two combinations that trailed a dedicated FM at 10 s were numerically superior with the full window, and the MAE reductions on PPV and SVV increased from 13.0\% and 16.6\% to 15.7\% and 24.9\%, respectively (Fig.~\ref{fig:performance}c). The pattern across tasks indicates the same tendency: the advantage was small on tasks that can be read from the morphology of a single beat, and largest on PPV and SVV, which require reading beat-to-beat variation across the respiratory cycle.

Integrating ten signals into one model did not degrade per-signal quality and, in at least one case, improved it. Given ECG or PPG alone, CARMEN matched or exceeded the signal-specific FMs in most comparisons (Table~\ref{tab:fm}; Fig.~\ref{fig:performance}b). The same held against FMs built to take multiple signals. NormWear \cite{NormWear}, built for multivariate wearable signals, and SPOTR \cite{SPOTR}, built for physiological signals in general, underperformed CARMEN at the same ECG+PPG input. Both encode channels independently or are pretrained on short fixed-length segments, and therefore appear not to capture cross-signal coupling and longer temporal structure in their representations. The gain on PPV has a plausible physiological mechanism. During mechanical ventilation, the pulse amplitude of arterial pressure and plethysmographic waveforms rises and falls with the respiratory cycle as intrathoracic pressure changes. In Phase 2, the airway pressure, respiratory flow, and capnogram of the same patient are input together with the hemodynamic signals, enabling the model to associate the respiratory variation in hemodynamic waveforms with the respiratory signals themselves (Fig.~\ref{fig:model}b, c). Consistent with this, completing Phase 2 reduced the PPV MAE by 8.6\% relative to the Phase 1 checkpoint (Fig.~\ref{fig:ablation}b).

Preserving absolute scale matters because clinicians judge by numbers as well as by waveform shape. Thresholds such as a MAP of 65 mmHg or an ICP of 20 mmHg decide whether to intervene, yet standard normalization removes exactly this information from the input. LSCNorm retains the training stability of normalization and returns the removed mean and standard deviation as conditioning (Fig.~\ref{fig:model}a). Without it, performance fell significantly on the two tasks in which an absolute value is the decision criterion (Intraoperative Hypotension and \etco{} Derangement), and the model greatly underestimated the risk of windows in which pressure was already low (Fig.~\ref{fig:ablation}a). The absolute level is not a detail that a large model automatically recovers; it has to be given back.

On two tasks with many labeled windows, training from scratch nearly closed the gap to CARMEN: on Sleep Apnea and Intraoperative Hypotension, each with hundreds of thousands of labeled segments, CARMEN-scratch came within 0.01 AUROC of CARMEN (Fig.~\ref{fig:performance}d), in line with reports from other domains \cite{He2019}. Sleep Apnea is also recorded with nasal airflow and respiratory effort belts in a sleep laboratory, a measurement setting absent from the pretraining cohorts. Hypoxemia and Intracranial Hypertension did not fit this pattern: both have few patients, yet CARMEN did not differ significantly from CARMEN-scratch (on Hypoxemia CARMEN-scratch was higher by 0.023). In cohorts this small the confidence intervals are wide, and these two comparisons can neither establish nor exclude a benefit of pretraining. The results by model size showed a related pattern. Of the three sizes, Large was best on physiological estimation and on tasks with large samples (8 of the 19 tasks), whereas Small was best on tasks with few patients such as Hypoxemia, In-Unit Mortality and Mechanical Ventilation (5 tasks; Supplementary Tables \hyperref[tab:s-detection]{S2}--\hyperref[tab:s-phenotype]{S6}). With small samples, the representation of a smaller model appears less prone to overfitting; Base was stable across all categories between these two tendencies and was therefore used as the primary model.

Preserving performance when channels are missing is a property of the training data more than of the architecture. Models with fixed inputs handle missingness by imputation or exclusion and therefore never see it during training. Most of CARMEN's training cohort comprised patients with only a subset of channels recorded, and the model was trained directly on that distribution. This is why a probe trained on all channels lost little when applied to patients with a channel missing (Fig.~\ref{fig:realworld}a): training on real ICU data is itself preparation for the deployment setting. In the other tasks as well, performance fell markedly only when the physiological signal that defines the task was removed, but held when the remaining channels were removed (Fig.~\ref{fig:realworld}b). This indicates that the representation depends on the task-relevant signals rather than on the number of channels.

This study has several limitations. First, although the pretraining corpus includes CVP, PAP, and ICP, only two downstream tasks take these signals as input, Cardiac Output and Intracranial Hypertension, and both have few patients (Fig.~\ref{fig:overview}c), because datasets with both invasive waveforms and clinical labels are rare. Second, external validation is limited in scope: the pretraining data come from a single institution, the Korean downstream cohorts are limited to the OR, and most US ICU tasks derive from a single database (MIMIC-III) collected in 2001--2012. Third, the labels of some tasks derive from nursing records or diagnosis codes and may contain noise and timing errors. These three limitations share a cause: the scarcity of labeled invasive-waveform data, and that scarcity is precisely what a representation learned without labels is meant to work around. Fourth, all downstream results are from a frozen encoder with a linear probe; fine-tuning may raise absolute performance but was outside the scope of this study, which concerns the quality of the pretrained representation. Fifth, the admission outcome tasks are the weakest part of the evaluation: CARMEN reached an AUROC of only 0.63--0.66 on Cardiac Arrest and Mechanical Ventilation (Supplementary Table \hyperref[tab:s-outcome]{S4}), reflecting the limits of predicting events over the following day from a few hours of waveform alone. Whether the waveform representation adds to the laboratory results and vital signs read alongside it in the ICU remains to be established.

Biosignal FMs initially employed ECG and PPG not because of clinical need but because public data were abundant for those signals. This study started from the opposite direction, collecting data in the OR and ICU, where the signals clinicians watch most closely are recorded. We let a single model read ten cardiorespiratory waveforms, from the electrical activity of the heart through arterial, venous, pulmonary artery, and intracranial pressures to respiratory flow, pressure, and gas exchange, while preserving their absolute values and combinations. We then showed that this frozen representation can be used across diverse OR and ICU tasks. To date, OR and ICU monitor waveforms have been read signal-by-signal and task-by-task. We hope that CARMEN can serve as a starting point for an approach in which a single model reads these waveforms and each new task is built on top of it.


\section{Methods}\label{sec:methods}

\subsection{Data and downstream tasks}\label{sec:meth-data}

\textbf{Pretraining cohorts.} Three continuous-waveform cohorts from SNUH were used for pretraining (Supplementary Table \hyperref[tab:datasets]{S1}). The SNUH OR and ICU cohorts were collected from bedside monitors and anesthesia machines in all ORs and four ICUs (one cardiopulmonary and three surgical) between January 2025 and September 2026, and comprised 75,567 records (527,773 modality-hours) and 256,346 records (1,078,097 modality-hours), respectively. K-MIMIC-MORTAL \cite{KMIMIC} comprises 194,148 records (668,831 modality-hours) of 725 adults who died during ICU admission between December 2023 and January 2025 and includes ICP, which was absent from the SNUH cohorts; electronic health records from K-MIMIC-MORTAL were not used. This deceased-patient cohort was included so that pretraining covered the physiological deterioration of the later ICU course and was used exclusively for pretraining.

\textbf{Downstream datasets and tasks.} Downstream tasks were constructed around clinical decisions made in the OR and ICU. Nineteen tasks were derived from eight datasets that were not used in pretraining (Fig.~\ref{fig:overview}c and Supplementary Table \hyperref[tab:datasets]{S1}). Data came from four settings: (1) SNUH ORs (VitalDB Open \cite{VitalDB}, VitalDB Arrhythmia Database \cite{VitalDBArrhythmia}, and the SNUH massive transfusion [MT] \cite{Lee2022MT, Kwon2024MT} and Ventilation cohorts), (2) US ICUs (Medical Information Mart for Intensive Care III Waveform Database [MIMIC-III WDB] \cite{MIMICIIIWDB} and MIMIC-III-Ext-PPG \cite{MIMICExtPPG}), (3) a sleep laboratory (CinC 2018 \cite{CinC2018}), and (4) a wearable device (Aurora-BP \cite{AuroraBP}). Clinical event detection (five tasks) identifies immediate abnormalities: arrhythmia, atrial fibrillation, hypoxemia, sleep apnea, and low cardiac output. Short-term event prediction (four tasks) predicts events that must be prepared for minutes in advance: intraoperative hypotension, massive transfusion, intracranial hypertension, and \etco{} derangement. Admission outcome prediction (four tasks) predicts admission-level outcomes from waveforms within the first 24 h of ICU admission: cardiac arrest, length of stay, initiation of mechanical ventilation, and death. Physiological estimation (four tasks) estimates from non-invasive signals the PPV, SVV, and MAP that otherwise require invasive arterial monitoring or dedicated equipment, together with the respiratory rate. Phenotype (two tasks: age and sex) checks whether patient characteristics are captured in the representation. In the prediction tasks, windows already in the event state at the end of the observation period were excluded. In estimation tasks, the signal used to derive the label was excluded from the input. Label definitions, inputs, and inclusion and exclusion criteria for each task are provided in \hyperref[note:tasks]{Supplementary Note 1}.

\textbf{Ethics.} This study was approved by the Institutional Review Board (IRB) of SNUH under two protocols (No.\ 2003-067-1108 and No.\ E-2608-020-1766). The requirement for informed consent was waived because retrospectively collected data were used. The public datasets received institutional ethics approval in their original studies and were used in accordance with the data use conditions of their distributors.

\begin{figure}[p]
\centering
\makebox[\textwidth][c]{%
\includegraphics[height=\dimexpr\paperheight-50mm\relax,width=\dimexpr\paperwidth-60mm\relax,keepaspectratio]{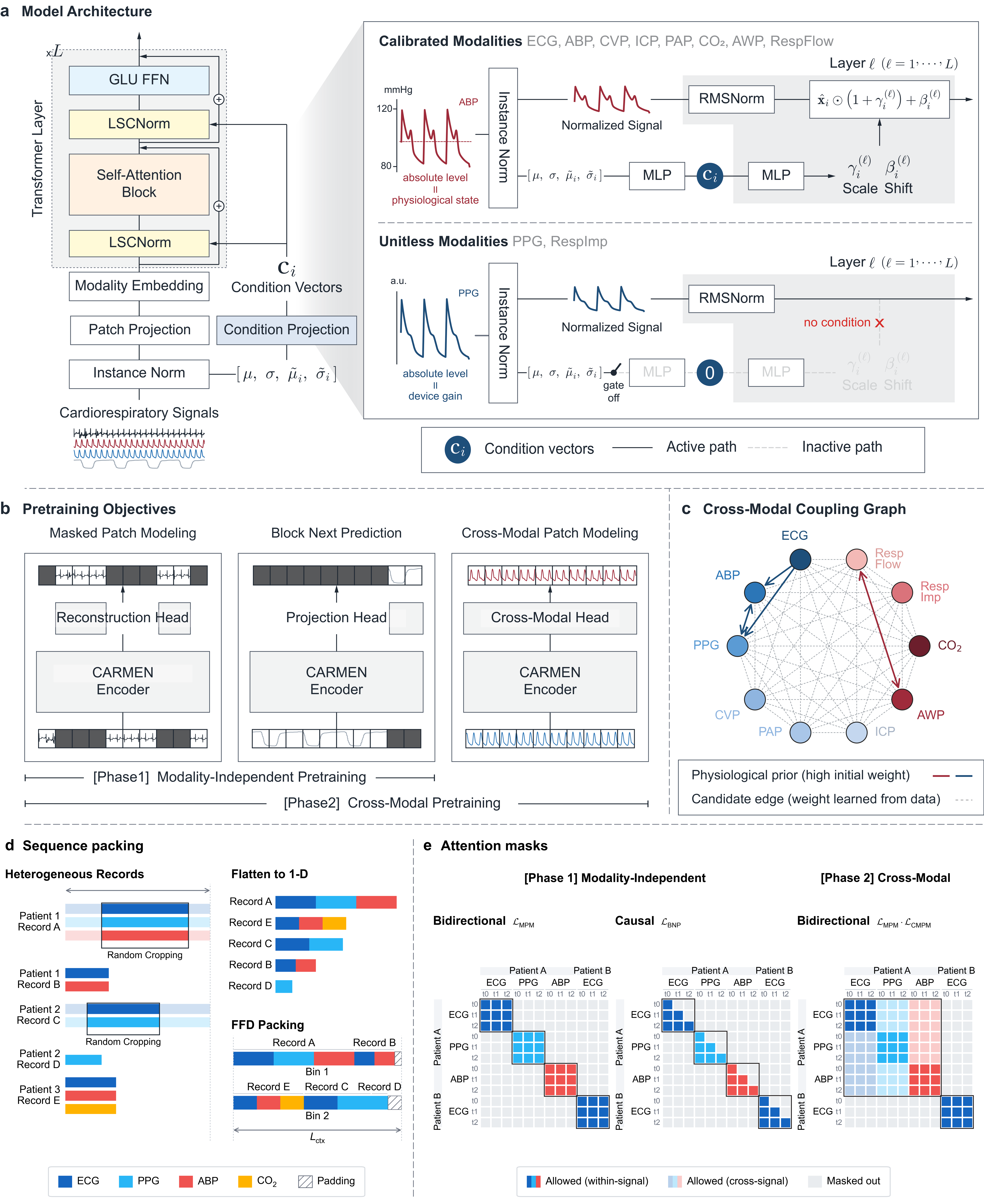}}
\caption{\textbf{Model architecture and pretraining of CARMEN.} Legend on the following page.}
\label{fig:model}
\end{figure}

\begin{figure}[!t]
\ContinuedFloat
\caption{\textbf{Model architecture and pretraining of CARMEN} (continued). \textbf{a}, Encoder and LSCNorm. The segment mean and standard deviation ($\mu$, $\sigma$) and the patch-level relative values ($\tilde{\mu}_i$, $\tilde{\sigma}_i$) form the conditioning vector $\mathbf{c}_i$, which sets the scale and shift in every layer. For PPG and RespImp, which have no absolute units, this path is switched off. \textbf{b}, Pretraining objectives: masked patch modeling, block next prediction and, in Phase 2, cross-modal patch modeling. \textbf{c}, Cross-modal coupling graph. Edge weights are learned; physiologically established pairs (solid lines) start from high values. \textbf{d}, Sequence packing. Unrecorded signals produce no tokens, long records are randomly cropped, and signals are packed by first-fit decreasing into sequences of the context length $L_{\mathrm{ctx}}$; only the remainder is padded. \textbf{e}, Attention masks for two patients. In Phase 1, each signal attends only to itself; in Phase 2, the co-recorded signals of a patient also attend to each other. Attention between patients is always blocked. LSCNorm, Loc/Scale-conditioned Layer Normalization; signal abbreviations as in Fig.~\ref{fig:overview}.}
\end{figure}
\subsection{Model architecture}\label{sec:meth-arch}

CARMEN is a patch-based single encoder that processes heterogeneous cardiorespiratory biosignals in one representation space. The design targets three problems. The first is large variation in amplitude distribution across patients and sensors. The second is semantic heterogeneity across modalities. The third is capturing local waveform dynamics together with long-range temporal dependencies. The overall structure is shown in Fig.~\ref{fig:model}a--c.

\subsubsection{Tokenization}

All signals are resampled to 100 Hz and divided into non-overlapping patches of $P=25$ samples (0.25 s). One beat (approximately 1 s) is split into about four patches. Intra-beat morphology, such as the dicrotic notch, is therefore resolved at the patch level. Respiratory signals (period 3--5 s) share the same temporal grid, so temporal alignment across modalities is preserved. Each patch is projected to a $\dmodel$-dimensional token by a residual multilayer perceptron (MLP) shared across all modalities \cite{TimesFM}. A modality embedding learned per modality is added to distinguish the signal type.

\subsubsection{Loc/Scale-conditioned Layer Normalization (LSCNorm)}

Biosignals differ in units and ranges. Without normalization, the model is biased toward modalities with large amplitude. The input segment of each modality channel is therefore standardized to zero mean and unit standard deviation by instance normalization at the input stage. However, for signals with physical units such as ABP, the absolute value itself is the clinical decision criterion (for example, MAP $<$ 65 mmHg or plateau pressure $>$ 30 cmH$_2$O). The mean and standard deviation removed by normalization are then precisely the key information.

Previous work has compensated for this loss by feeding the removed mean and standard deviation back into the model. Reversible instance normalization (RevIN) \cite{RevIN} restores them at the output. Non-stationary Transformers \cite{NSTransformer} incorporate them into the attention logits. LSCNorm, proposed here, converts the mean and standard deviation into a conditioning vector and injects it into the normalization of every Transformer layer. LSCNorm replaces only RMSNorm. The attention and feed-forward network (FFN) structures are unchanged.

One conditioning vector is produced per patch. It is built from the mean and standard deviation of the whole input segment and of that patch. The segment-level mean $\mu$ and standard deviation $\sigma$ are kept in physical units and represent the absolute level of the signal. The mean $\mu_i$ and standard deviation $\sigma_i$ of patch $i$ are expressed relative to the segment-level values and represent temporal change within the segment:
\begin{equation}
\tilde{\mu}_i=\frac{\mu_i-\mu}{\sigma},\qquad \tilde{\sigma}_i=\frac{\sigma_i}{\sigma},\qquad \mathbf{c}_i=\mathrm{MLP}\bigl([\,\mu,\ \sigma,\ \tilde{\mu}_i,\ \tilde{\sigma}_i\,]\bigr).
\end{equation}
Each layer $\ell$ transforms the conditioning vector $\mathbf{c}_i$ into a per-dimension scale and shift $(\gamma_i^{(\ell)},\beta_i^{(\ell)})$ and applies them to the RMSNorm output. This is affine conditioning, as applied in feature-wise linear modulation (FiLM) \cite{FiLM}, restricted to the normalization layer:
\begin{equation}
\begin{gathered}
\hat{\mathbf{x}}_i = \mathrm{RMSNorm}(\mathbf{x}_i), \qquad
[\gamma_i^{(\ell)};\,\beta_i^{(\ell)}] = \mathrm{Linear}^{(\ell)}(\mathbf{c}_i),\\
\mathrm{LSCNorm}^{(\ell)}(\mathbf{x}_i;\mathbf{c}_i) = \hat{\mathbf{x}}_i \odot \bigl(1+\gamma_i^{(\ell)}\bigr)+\beta_i^{(\ell)}.
\end{gathered}
\end{equation}
The MLP (two linear layers with a 16-unit hidden layer and SiLU activation) is shared across modalities and receives no modality identifier; $\mu$ and $\sigma$ enter in physical units without rescaling.
$\mathrm{Linear}^{(\ell)}$ is initialized to zero. LSCNorm is therefore identical to standard RMSNorm at the start of training. This design, with the scale set to $(1+\gamma)$, follows the adaptive layer normalization with zero initialization (adaLN-Zero) of the Diffusion Transformer (DiT) \cite{DiT} and prevents abrupt early changes in the representation due to conditioning.

\textbf{Modality-selective gating.} The meaning of the mean and standard deviation differs by modality. For signals with absolute units, such as ECG, ABP, CVP, PAP, ICP, \co, AWP, and RespFlow, the mean and standard deviation carry physiological information. In contrast, the amplitude and baseline of PPG are set by the device's automatic gain control. RespImp is an impedance change on an arbitrary scale. For these two signals, the values may reflect device differences rather than physiology. For PPG and RespImp tokens, we therefore set $\mathbf{c}_i=\mathbf{0}$ to block the conditioning path. LSCNorm then reduces to an input-independent affine normalization in which only the bias of $\mathrm{Linear}^{(\ell)}$ remains.

\subsubsection{Transformer backbone}

The token sequence is processed by a pre-norm Transformer encoder of $L$ layers. Each layer consists of multi-head attention and a gated linear unit (GLU) FFN \cite{Shazeer2020GLU}. The normalizations immediately before the two sub-blocks and at the encoder output are all LSCNorm. Statistical conditioning is thus injected at every layer.

Position is encoded with rotary position embedding (RoPE) \cite{RoPE}. The clinical meaning of biosignals arises from relative timing between points: inter-beat intervals, the delay between the peak and the dicrotic notch, and the phase relationship between heartbeat and respiration. RoPE expresses these relative distances directly within attention. It has no learned position table: the rotation angles are computed from the position index with fixed frequencies, so the encoder has no maximum input length. The same weights therefore process the 10-s to 5-min segments of pretraining (40 to 1,200 patches per signal) and longer downstream windows, up to the 10 min (2,400 patches per signal) of Massive Transfusion and Intracranial Hypertension, without modification. Attention is masked to prevent crossing sequence boundaries. This mask also determines which signals reference each other within a single forward pass (Fig.~\ref{fig:model}e).

Three model sizes are used: Small (52M; $\dmodel=512$, $L=16$, $h=8$), Base (174M; $\dmodel=768$, $L=24$, $h=12$), and Large (410M; $\dmodel=1024$, $L=32$, $h=16$). The three models were trained on the same data, with the same training settings and the same number of tokens.

\subsection{Pretraining}\label{sec:meth-pretrain}

\subsubsection{Data preparation and augmentation}

\textbf{Sequence packing and random cropping.} The combination of observed modalities and recording length differs from patient to patient. CARMEN composes its input from the observed signals only, so missing modalities need not be imputed. Records were divided into 5-min units. A segment at a random position was then cropped with a ratio $\rho\sim\mathrm{Uniform}(0.01,1.0)$ of the unit length, with the cropped length floored at 40 patches (10 s), so that the input length varied between 10 s and 5 min; records shorter than 10 s were used at their full length. The patch tokens of a channel's cropped segment, joined together, form a sequence. Sequences of different lengths were packed into a fixed-length context (Supplementary Table \hyperref[tab:s-hyper]{S7}) by first-fit decreasing \cite{Krell2021} to minimize padding (Fig.~\ref{fig:model}d). Attention was restricted to each sequence, and the RoPE position index was reset to 0 for each sequence. Therefore, no information mixed between unrelated records placed in the same context. The attention scope was a single signal in Phase 1 and the set of signals recorded simultaneously from the same patient in Phase 2. In downstream evaluation, the channels of a patient were likewise encoded together as one sequence (Fig.~\ref{fig:model}e).

\textbf{Modality-balanced sampling.} ECG and PPG are measured in most records, whereas invasive signals such as CVP and PAP are rare. Training on the raw distribution gives rare modalities too few learning opportunities. To mitigate this, we applied temperature-based sampling \cite{Lample2019}, which rescales sampling probability according to modality frequency. To prevent oversampling of a small number of records, the sampling weight was capped at five times the overall mean.

\textbf{Amplitude jitter.} Gaussian noise proportional to the standard deviation of each channel segment (0.02 for ECG, 0.05 otherwise) was added. The model thus learns a representation that relies on waveform shape rather than on fine amplitude changes.

\subsubsection{Curriculum learning}

CARMEN follows a two-phase curriculum \cite{Bengio2009Curriculum}. It first learns intra-modality representations and then learns cross-modality coupling. If, from the outset, several modalities are trained together, the model may exploit the correlation between signals synchronized to the cardiac cycle as a shortcut before learning each signal's morphological characteristics \cite{Geirhos2020Shortcut}. This phenomenon parallels greedy learning, in which the fastest-converging modality dominates optimization and the others are undertrained \cite{Wu2022Greedy}.

\textbf{Phase 1: Modality-independent pretraining.} Each sample is one modality segment from one patient. A batch mixes multiple patients and modalities. Since attention is blocked at sample boundaries, there is no information exchange between signals (Fig.~\ref{fig:model}e, Phase 1). With no other signal available as a cue, the model must reconstruct the masked and subsequent segments from each signal's own waveform morphology and temporal variation alone. Only intra-modal objectives are optimized in this phase.

\textbf{Phase 2: Cross-modal pretraining.} Training starts from the Phase 1 weights. Multiple modalities recorded simultaneously from one patient are grouped into one sample. Within a sample, attention opens across signal boundaries. Different signals at the same time point can therefore reference each other directly (Fig.~\ref{fig:model}e, Phase 2). In this phase, the cross-modal objective is added to the intra-modal objectives. The intra-modal objectives are retained so that the per-signal representations learned in Phase 1 do not weaken while cross-signal coupling is being learned.

\subsubsection{Objective functions}

Phase 1 optimizes masked patch modeling (MPM) and block next prediction (BNP). Phase 2 adds cross-modal patch modeling (CMPM). All three objectives share the same patch-level loss $\ell_{\mathrm{patch}}$.

\textbf{Intra-modal objectives.} MPM masks 40\% of all patches in contiguous blocks of 24--64 patches (6--16 s) and reconstructs them. Masking single patches would allow reconstruction by local interpolation from the neighboring patches alone, so blocks were masked instead. The 6-s lower bound was chosen so that one full respiratory cycle (approximately 5 s) fits within a block. BNP predicts the next $K=24$ patches (6 s) from a causal pass in which future references are blocked. It does so in a single forward pass, without autoregressive iteration. The encoder is called twice per step: MPM uses the output of the bidirectional pass, and BNP that of the causal pass.
Block lengths are capped by the remaining masking budget, so a segment shorter than 60 patches (15 s) receives a single block shorter than 24 patches (16 patches for a 40-patch segment). BNP targets that fall beyond the end of the segment are excluded from the loss.
 Let $\mathcal{M}$ be the set of masked patches, $\mathbf{h}_i$ the encoder output at position $i$, $\mathbf{p}_i\in\mathbb{R}^{P}$ the original patch, $f_{\mathrm{recon}}$ the reconstruction head, and $g$ and $f_k$ the shared BNP trunk (Linear--GELU, where GELU is the Gaussian error linear unit) and the $k$-th prediction head; then
\begin{equation}
\mathcal{L}_{\mathrm{MPM}} = \frac{1}{|\mathcal{M}|}\sum_{i\in\mathcal{M}}\ell_{\mathrm{patch}}\bigl(f_{\mathrm{recon}}(\mathbf{h}_i),\,\mathbf{p}_i\bigr),\qquad
\mathcal{L}_{\mathrm{BNP}} = \mathbb{E}_i\!\left[\frac{1}{K}\sum_{k=1}^{K}\ell_{\mathrm{patch}}\bigl(f_k(g(\mathbf{h}_i)),\,\mathbf{p}_{i+k}\bigr)\right].
\end{equation}

\textbf{Cross-modal objective.} CMPM reconstructs a masked patch $j$ from patch $i$ of another modality observed at the same time point. If the source is masked, reconstruction becomes contextual inference within the same modality. If the target is unmasked, the original remains in the input and a copying shortcut arises. Only pairs with an observed source and a masked target were therefore used. Let $\hat{\mathbf{p}}_{i\to j}$ be the target-modality output of the cross-modal head $f_{\mathrm{cross}}$ and $\mathcal{P}_{s\to t}$ the set of patch pairs belonging to the ordered modality pair $(s,t)$; then
\begin{equation}
\ell_{s\to t}=\frac{1}{|\mathcal{P}_{s\to t}|}\sum_{(i,j)\in\mathcal{P}_{s\to t}}\ell_{\mathrm{patch}}\bigl(\hat{\mathbf{p}}_{i\to j},\,\mathbf{p}_j\bigr),\qquad
\mathcal{L}_{\mathrm{CMPM}}=\frac{\sum_{s\neq t}W_{s\to t}\,\ell_{s\to t}}{\sum_{s\neq t}W_{s\to t}}.
\end{equation}
Averaging within each pair first prevents pairs with long co-measurement time from dominating the loss. The coupling weight is $W_{s\to t}=\sigma(A_{s\to t})$, where $A$ is learned together with the objective. Physiologically well-established pairs start from $\sigma(2)\approx0.88$: ECG$\to$ABP \cite{Bers2002}, ECG$\to$PPG \cite{Mukkamala2015}, ABP$\leftrightarrow$PPG \cite{Allen2007}, and AWP$\leftrightarrow$RespFlow \cite{Bates2009}. The remaining pairs start from $\sigma(-2)\approx0.12$. The learned $W$ weights the modality pairs in the objective. Under this normalized form, a more easily reconstructed pair receives a larger weight. Consequently, $W$ may reflect reconstruction difficulty as well as physiological coupling, and we do not interpret it here.

\textbf{Patch-level loss and total objective.} $\ell_{\mathrm{patch}}$ is the sum of an amplitude-weighted mean squared error (MSE) and a multi-resolution short-time Fourier transform (STFT) loss. A per-sample weight $w_t=1+|p_t|/\max_{t'}|p_{t'}|$ was applied. This prevents flat baselines from dominating the loss and weakening the reconstruction of high-amplitude regions such as beat peaks. $\ell_{\mathrm{STFT}}$ was added because point-wise MSE favors over-smoothed reconstructions and under-penalizes errors in sharp transients such as the QRS upstroke and the dicrotic notch. It is the mean of spectral convergence and log-magnitude $L_1$ error over STFTs with window sizes 8 and 16 samples (80 and 160 ms; hop one quarter of the window). At these window lengths, the term constrains the local spectral envelope within a patch; beat-level periodicity is instead learned through the multi-second masking and next-prediction objectives. The final objective includes an L1 term on the mean coupling weight $\overline{W}$:
\begin{equation}
\begin{aligned}
\ell_{\mathrm{patch}}(\hat{\mathbf{p}},\mathbf{p})&=\frac{1}{P}\sum_{t=1}^{P}w_t(\hat p_t-p_t)^2+0.2\,\ell_{\mathrm{STFT}}(\hat{\mathbf{p}},\mathbf{p}),\\
\mathcal{L}&=\mathcal{L}_{\mathrm{MPM}}+0.8\,\mathcal{L}_{\mathrm{BNP}}+0.3\,\mathcal{L}_{\mathrm{CMPM}}+0.005\,\overline{W}.
\end{aligned}
\end{equation}
Without the L1 term, all coupling weights converge to 1, and the model does not select pairs.
 The last two terms are not used in Phase 1.

\subsection{Evaluation protocol}\label{sec:meth-eval}

\textbf{Data separation and split.} The ten tasks derived from external institutions cannot overlap with the pretraining records because the institutions differ. The nine tasks from the same institution used OR recordings collected between 2016 and December 2024. The SNUH OR and ICU pretraining cohorts were collected from January 2025 onward, and K-MIMIC-MORTAL (December 2023 to January 2025) contains ICU recordings only, so no recording appears in both pretraining and evaluation. Within each task, patient-level 5-fold cross-validation was performed, and performance was computed over all out-of-fold predictions. The evaluation unit depends on the task category. It is the window for clinical event detection, short-term event prediction, and physiological estimation; the patient for phenotype; and the admission for admission outcome prediction. For the latter two, window-level predictions or embeddings were averaged.

\textbf{Linear probing.} All pretrained models were evaluated with a linear probe on a frozen encoder. This isolates the quality of the learned representation from task-specific fine-tuning effects. The encoder was frozen at its pretrained weights, and only the task-specific head was trained. The outputs were mean-pooled for each channel. For multimodal inputs, per-channel embeddings were concatenated in a fixed order. The head consisted of layer normalization, dropout, and a single linear layer. It was trained with Adam using cross-entropy for classification and MSE for regression (Supplementary Table \hyperref[tab:s-hyper]{S7}).

\textbf{Baselines.} Two supervised comparators were used: 1D-ResNet and CARMEN-scratch. The latter has the same architecture as CARMEN but was trained from a random initialization. Both received 100-Hz signals with as many channels as the input modalities of the task. As pretrained baselines, MOMENT \cite{MOMENT}, a general-purpose time-series FM, and 11 biosignal FMs were compared using the same linear probe. For admission outcome prediction, only pretrained models were compared using a linear probe, without supervised comparators.

The 11 biosignal FMs comprise (1) five ECG FMs (ST-MEM \cite{STMEM}, ECG-JEPA \cite{ECGJEPA}, HuBERT-ECG \cite{HuBERTECG}, ECGFM-KED \cite{ECGFMKED}, and MERL \cite{MERL}), (2) three PPG FMs (AnyPPG \cite{AnyPPG}, PaPaGei \cite{PaPaGei}, and Pulse-PPG \cite{PulsePPG}), and (3) three FMs that take ECG and PPG together (BIOT \cite{BIOT}, NormWear \cite{NormWear}, and SPOTR \cite{SPOTR}). They were compared for the six ECG and PPG tasks (Table~\ref{tab:fm}) with the pretrained weights released by their authors. Each received an input of its own pretraining length, with CARMEN's input modality matched to it, and the 12-lead ECG FMs received the single available lead replicated across the 12 leads. NormWear and SPOTR were also evaluated for all 19 tasks using the same pipeline as CARMEN, with only the encoder exchanged (Fig.~\ref{fig:performance} and Supplementary Tables \hyperref[tab:s-detection]{S2}--\hyperref[tab:s-phenotype]{S6}).

\textbf{Metrics.} The primary metrics were AUROC for classification and MAE for regression. The areas under the precision--recall curve (AUPRC) and Pearson $r$ are reported in Supplementary Tables \hyperref[tab:s-detection]{S2}--\hyperref[tab:s-phenotype]{S6}. The 95\% confidence intervals (CIs) were obtained using patient-cluster bootstrapping (2,000 iterations). The differences between the models $\Delta$ were obtained using a paired bootstrap that shared the same resamples. For multiple comparisons, Benjamini--Hochberg correction was applied per metric within each Results subsection. Statistical significance was defined at $q<0.05$.

\textbf{Ablation setting.} Because changing LSCNorm or the patch size requires pretraining from scratch, the ablations were run at smaller training budgets than CARMEN-Base, and each variant was compared with a reference model trained with the same data, architecture, budget and learning-rate schedule. Token counts are given as packed context length and include padding. For LSCNorm (Fig.~\ref{fig:ablation}a) and Phase 2 (Fig.~\ref{fig:ablation}b), the reference model was trained on about 59 billion tokens, about half the budget of CARMEN-Base, and the Phase 2 comparison used the Phase 1 checkpoint of this model. Patch size (Fig.~\ref{fig:ablation}c) was evaluated in a reduced setting with fewer steps per epoch in both phases (about 8.9 billion tokens, 15\% of the reference model), and within this setting the 0.25 s patch model is the reference.


\section*{Supplementary information}

\hyperref[note:tasks]{Supplementary Note 1} and Supplementary Tables \hyperref[tab:datasets]{S1}--\hyperref[tab:s-hyper]{S7} follow the main text.

\section*{Acknowledgments}

This research was supported by a grant from the Korea Health Technology R\&D Project through the Korea Health Industry Development Institute (KHIDI), funded by the Ministry of Health \& Welfare, Republic of Korea (grant number: RS-2024-00439677, NTIS number: 2460003917).

\section*{Declarations}

\subsection*{Data availability}
The public datasets used for the downstream evaluations are accessible from their respective distributors. VitalDB Open \cite{VitalDB}, the VitalDB Arrhythmia Database \cite{VitalDBArrhythmia}, the MIMIC-III Waveform Database \cite{MIMICIIIWDB}, and the CinC 2018 database \cite{CinC2018} are openly available on PhysioNet; MIMIC-III-Ext-PPG \cite{MIMICExtPPG} is available through PhysioNet credentialed access; and Aurora-BP \cite{AuroraBP} is available on Zenodo after execution of a data use agreement. K-MIMIC-MORTAL \cite{KMIMIC} is provided on a restricted basis to researchers approved through the Korea Health Data Platform (KHDP) operated by SNUH. The SNUH OR, ICU, MT, and Ventilation cohorts are not publicly available to protect patient privacy. Researchers with approval from the SNUH IRB may inquire about access through a reasonable request to the corresponding author.

\subsection*{Code availability}
The CARMEN implementation is available on GitHub (\url{https://github.com/dlcjfgmlnasa/CARMEN}). The pretrained weights of the three sizes (Small 52M, Base 174M, and Large 410M; files \texttt{carmen-small.pt}, \texttt{carmen-base.pt}, and \texttt{carmen-large.pt}) are on the Hugging Face Hub (\url{https://huggingface.co/dlcjfgmlnasa28/CARMEN}). The code and weights are released under the PolyForm Noncommercial License 1.0.0, which permits research and other noncommercial uses. The modules ported from uni2ts \cite{Moirai} (\url{https://github.com/SalesforceAIResearch/uni2ts}), which cover part of the packed-batch normalization and attention, remain under the Apache License 2.0. The code was implemented in PyTorch, and the reported results were obtained using the Base model.
The comparison models used the implementations and pretrained weights released by their respective authors without modification: MOMENT \cite{MOMENT}, \url{https://github.com/moment-timeseries-foundation-model/moment}; NormWear \cite{NormWear}, \url{https://github.com/Mobile-Sensing-and-UbiComp-Laboratory/NormWear}; SPOTR \cite{SPOTR}, \url{https://github.com/5GYYYYY/SPOTR}; BIOT \cite{BIOT}, \url{https://github.com/ycq091044/BIOT}; ST-MEM \cite{STMEM}, \url{https://github.com/vuno/ST-MEM}; ECG-JEPA \cite{ECGJEPA}, \url{https://github.com/sehunfromdaegu/ECG_JEPA}; HuBERT-ECG \cite{HuBERTECG}, \url{https://github.com/Edoar-do/HuBERT-ECG}; ECGFM-KED \cite{ECGFMKED}, \url{https://github.com/control-spiderman/ECGFM-KED}; MERL \cite{MERL}, \url{https://github.com/cheliu-computation/MERL-ICML2024}; AnyPPG \cite{AnyPPG}, \url{https://github.com/PKUDigitalHealth/AnyPPG}; PaPaGei \cite{PaPaGei}, \url{https://github.com/Nokia-Bell-Labs/papagei-foundation-model}; Pulse-PPG \cite{PulsePPG}, \url{https://github.com/maxxu05/pulseppg}. Training and evaluation were performed using Python 3.13 and PyTorch 2.10.

\subsection*{Author contributions}
C.-H.L.\ conceived the study, developed the model, performed the experiments, and wrote the manuscript. W.K.\ and H.L.\ contributed to writing the manuscript. G.L.\ and D.-I.E.\ provided the SNUH MT and Ventilation cohorts, respectively. H.-C.L.\ supervised the study, acquired the data, and revised the manuscript. All the authors have read and approved the final version.

\subsection*{Competing interests}
H.L.\ is an associate editor at \textit{npj Digital Medicine}; all other authors declare no competing interests.

\clearpage
\setcounter{table}{0}
\renewcommand{\thetable}{S\arabic{table}}

\section*{Supplementary Information}

\subsection*{Supplementary Note 1. Downstream task definitions}\label{note:tasks}


The input window, observation period, horizon, analysis scale and prevalence of each task are summarized in Fig.~\ref{fig:overview}c of the main text.

\subsubsection*{Clinical event detection}

\textbf{Arrhythmia.} The arrhythmia type of 30-s segments was classified in 482 surgeries (473 patients) from VitalDB \cite{VitalDB} for which specialist-verified rhythm annotations are available \cite{VitalDBArrhythmia}. The original annotations were consolidated into four classes: normal sinus rhythm (NSR), atrial fibrillation/flutter (AF/AFL), ventricular arrhythmia (ventricular tachycardia and patterned ventricular ectopy), and bradycardia/conduction disorder (sinus node dysfunction and atrioventricular block). Inputs were ECG and PPG, and only segments in which both signals were valid were analyzed: 17,295 30-s segments from 468 patients had valid ECG, and 14,709 segments from 413 patients had both ECG and PPG.

\textbf{Atrial Fibrillation.} In MIMIC-III-Ext-PPG \cite{MIMICExtPPG}, the presence of atrial fibrillation or atrial flutter in 30-s segments was classified from PPG alone. Two hundred patients were selected for each of five rhythms: sinus rhythm (SR), atrial fibrillation (AF), sinus tachycardia (STACH), sinus bradycardia (SBRAD) and atrial flutter (AFLT) (rhythm abbreviations follow the original dataset). The 15 patients without any segment of valid ECG were excluded, and 985 patients were analyzed. Patients with multiple rhythms were assigned to a single class with priority given to the rarer rhythm (AFLT $>$ SBRAD $>$ AF $>$ STACH $>$ SR), and up to 20 segments of the assigned rhythm were randomly sampled per patient. AF and AFLT were taken as positive and SR, STACH and SBRAD as negative. Both rhythms are atrial tachyarrhythmias with essentially the same stroke risk and anticoagulation criteria, and they are difficult to distinguish from PPG alone.

\textbf{Hypoxemia.} Hypoxemia was classified in 188 mechanically ventilated surgeries from the SNUH anesthesia adverse-event records (SNUH Ventilation cohort). Labels were generated from the 1-Hz peripheral oxygen saturation (\spo) trend of the bedside monitor; a 30-s window was positive if any quality-filtered \spo{} value within it was below 90\%. To reduce spurious \spo{} drops due to motion or hypoperfusion, only samples satisfying \spo{} 50--100\%, a relative difference between pulse-derived and ECG-derived heart rate $\leq 20\%$, and perfusion index (PI) $\geq 0.3$ were used for label derivation. Only the three respiratory signals \co, AWP and RespFlow were input, and the 167 surgeries in which all required signals and the \spo{} trend were valid were analyzed.

\textbf{Sleep Apnea.} Polysomnography recordings of the PhysioNet/Computing in Cardiology Challenge 2018 (CinC 2018) \cite{CinC2018} were divided into non-overlapping 60-s windows, and a window overlapping an apnea or hypopnea event by 10 s or more was defined as positive. Inputs were nasal airflow and thoracic and abdominal respiratory effort belts, mapped to the closest pretraining modalities, RespFlow and RespImp, respectively.
Nasal airflow is recorded in arbitrary units, whereas the LSCNorm conditioning of RespFlow assumes physical units; the conditioning was left active, so the segment statistics supplied for this channel do not carry their pretraining meaning.

\textbf{Cardiac Output.} Low cardiac output was classified in ICU patients from MIMIC-III WDB \cite{MIMICIIIWDB} in whom a PAP waveform was recorded via a pulmonary artery catheter. A charted thermodilution cardiac output (CO) below 4.0 L/min was defined as positive. The input was the 60 s of PAP waveform immediately preceding each thermodilution measurement (or the 60 s immediately following if no preceding segment was available); CVP recorded in the same segment was input together when present and treated as a missing channel otherwise. CVP was co-recorded in 35.2\% of all windows. To prevent a few patients with many repeated measurements from dominating the results, up to 20 measurements were randomly sampled per patient.

\subsubsection*{Short-term event prediction}

\textbf{Intraoperative Hypotension.} Intraoperative hypotension was predicted in 2,764 surgeries (2,657 patients) from VitalDB \cite{VitalDB} with valid ABP recordings. Hypotension was defined as MAP below 65 mmHg sustained for 1 min or longer \cite{Salmasi2017, Hatib2018}. Five minutes of ABP were input to predict the onset of new hypotension within 5 min after the end of observation, and windows already hypotensive at the end of observation were excluded.

\textbf{Massive Transfusion.} Massive transfusion was predicted in 5,850 of the 17,986 surgeries under general anesthesia performed at SNUH between August 2016 and November 2019 in which ABP and PPG were recorded continuously \cite{Lee2022MT, Kwon2024MT}. Massive transfusion was defined as administration of 3 or more units of packed red blood cells within 1 h. Ten minutes of ABP and PPG were input to predict massive transfusion beginning 5--10 min after the end of observation. In positive surgeries the window ended 5 min before the onset of massive transfusion; if that window was not valid, it was moved earlier in 30-s steps by up to 5 min. In negative surgeries a 30-min span was drawn uniformly between the start of surgery and 30 min before its end, and one 10-min window with at least 80\% valid samples was drawn at random within it. Negative windows were not matched to positive windows on elapsed surgical time, which may therefore contribute to discrimination.

\textbf{Intracranial Hypertension.} Intracranial pressure crises were predicted in critically ill patients with ICP recordings in the MIMIC-III WDB matched subset \cite{MIMICIIIWDB}. A crisis was defined as the 1-min median ICP exceeding 20 mmHg for 5 min or longer \cite{Guiza2013}. Ten minutes of ABP and ICP were input to predict onset within 5 min after the end of observation, and windows in which ICP already exceeded 20 mmHg at the end of observation were excluded.

\textbf{\etco{} Derangement.} \etco{} derangement was predicted in 1,000 surgeries randomly sampled from the 6,338 surgeries (of the 6,388 in VitalDB Open \cite{VitalDB}) in which the anesthesia machine's \co{} and AWP waveforms and end-tidal \co{} (\etco) were all recorded. A window was positive if the mean \etco{} over the 1 min beginning 3 min after the end of observation was below 30 mmHg or above 45 mmHg, and windows already outside the 30--45 mmHg range at the end of observation were excluded. The input was 60 s of the \co{} and AWP waveforms only; the numeric \etco{} value reported by the monitor was not input.

\subsubsection*{Admission outcome prediction}

All four tasks use the MIMIC-III WDB matched subset \cite{MIMICIIIWDB}. The observation period was divided into non-overlapping windows, and in each window the recorded ECG, PPG, ABP and RespImp were encoded together as a single input sequence. The per-channel token means were averaged per patient and concatenated in channel order to form the admission-level representation (channels not recorded were given a zero vector and a missingness indicator). Because MOMENT takes a single signal, its representation of ECG alone was averaged per patient in the same way for comparison; for NormWear and SPOTR, the recorded channels were encoded separately and averaged, and the result was used in the same way. Unless stated otherwise, up to 36 windows of 5 min (3 h) were used. The input for each patient was every one of the four channels recorded during the observation period; missing channels were not imputed and no patient was excluded for missing channels. The proportion of patients with each channel recorded was 95.2--98.5\% for ECG, 55.3--65.6\% for PPG, 13.9--46.2\% for ABP and 46.1--51.9\% for RespImp (range across tasks; Cardiac Arrest based on the 24-h cohort), and 6.4--16.6\% of patients had all four channels recorded.

\textbf{Cardiac Arrest.} The occurrence of cardiac arrest during the ICU stay was predicted at the admission level. Cardiac arrest was defined as meeting one or more of the following: ICD-9 (International Classification of Diseases, 9th Revision) diagnosis code 427.5, procedure codes 99.60 and 99.63, the MetaVision Cardiac Arrest event (ITEMID 225466), or the clinical-note-based annotation of MIMIC-III-Ext-CA \cite{MIMICExtCA}. The observation period was the first 3 h after ICU admission, and occurrence within 24 h after the end of observation was positive. When the time of occurrence was not recorded, only patients with an ICU stay of 27 h or less were included as positive, and two patients with occurrence within the observation period were excluded. Of the 10,282 patients with waveforms, two non-arrest control patients matched on age, sex, ICU type and ICU length of stay were selected for each of the 532 cardiac-arrest patients, so the prevalence is a consequence of the design. Of these, 273 cardiac-arrest patients with waveforms in the first 3 h and their 546 matched controls (819 patients) entered preprocessing, 790 (259 positive) remained, and after applying the 24-h label criterion 108 positive patients and the controls of the same matched sets, 320 patients in total, were analyzed.

\textbf{ICU Length of Stay.} Whether the first ICU length of stay (LOS) exceeded 3 days was predicted \cite{Harutyunyan2019}. The observation period was the first 3 h after ICU admission. Patients who died in the ICU after day 3 were kept as positive, and patients with LOS $\leq$ 3 days who died were excluded to rule out short stays due to early death. Of 9,928 eligible adults, the 5,361 with waveforms recorded within 3 h of admission were preprocessed, and the 5,128 with valid windows were analyzed.

\textbf{Mechanical Ventilation.} Initiation of invasive mechanical ventilation was predicted 24 h in advance in adults whose first ICU stay was in a medical ICU (MICU) or coronary care unit (CCU). Surgical ICUs were excluded because of the likelihood of planned perioperative intubation. The time of ventilation onset was defined as the earlier of the first charted ventilator setting (ventilation mode, positive end-expiratory pressure [PEEP], tidal volume, airway pressure, etc.) and the start time of the MetaVision procedure record (ITEMID 225792) \cite{MIMICCode}. The time at which a waveform was first recorded within 12 h of ICU admission was set as landmark L, [L, L+1 h] was the observation period, and initiation of mechanical ventilation in (L+1 h, L+25 h] was positive. Patients already ventilated before the end of observation, or who died within the following 24 h without mechanical ventilation, were excluded. In this task up to 60 windows of 1 min were used.

\textbf{In-Unit Mortality.} Death during the first ICU stay was predicted with a 24-h landmark design. The observation period was 21--24 h after ICU admission, and death between 24 h after admission and 6 h after the first ICU discharge was positive. To prevent overlap between the observation period and the outcome, patients who died or were discharged within 24 h of admission were excluded, and the positive patients of the Cardiac Arrest task were also excluded so that the two outcomes are mutually exclusive.

\subsubsection*{Physiological estimation}

\textbf{Pulse Pressure Variation (PPV).} PPV was estimated from PPG in VitalDB Open \cite{VitalDB} surgeries in which ABP, PPG and AWP were recorded simultaneously. The label was defined as the median within a 60-s window of the PPV computed for each respiratory cycle from the beat-to-beat pulse pressure of ABP. Windows with arrhythmia, irregular breathing, apnea, or a heart-rate/respiratory-rate ratio $<$ 3.6 were excluded. ABP and AWP were used only for label derivation and only PPG was input; up to 40 windows were sampled per patient (228 patients, 229 surgeries).

\textbf{Stroke Volume Variation (SVV).} SVV was estimated from PPG in VitalDB Open surgeries in which SVV was recorded by EV1000 or Vigileo. The label was the mean device SVV in each 60-s window, with the same window selection and physiological exclusion criteria as PPV (161 patients, 163 surgeries).

\textbf{Mean Arterial Pressure (MAP).} MAP was estimated from PPG in VitalDB Open surgeries in which ABP and PPG were recorded simultaneously. The label was the ABP-derived mean MAP of each non-overlapping 30-s window. Only windows with a valid-signal proportion $\geq 80\%$, systolic blood pressure (SBP) 70--260 mmHg, diastolic blood pressure (DBP) 30--150 mmHg, pulse pressure $\geq 10$ mmHg and at least three beats were used.

\textbf{Respiratory Rate (RR).} Respiratory rate (RR) was estimated from ECG and PPG in ICU patients of the MIMIC-III WDB matched subset \cite{MIMICIIIWDB} in whom ECG, PPG and RespImp were recorded simultaneously. The label was the dominant frequency of RespImp in the 0.1--0.6 Hz band in each non-overlapping 60-s window. Windows in which the dominant component was below 25\% of the band power or the respiratory rate was outside 6--36 breaths/min were excluded, and only segments in which the valid proportion of each of the three signals was 95\% or more were used. RespImp was used only for label derivation, and up to 60 windows per patient were sampled uniformly across the full recording.

\subsubsection*{Phenotype}

\textbf{Age.} Age was estimated from wearable ECG (lead I) and PPG in the resting-posture laboratory measurements of Aurora-BP \cite{AuroraBP}. Each measurement was divided into 15-s windows and encoded, and window-level predictions were averaged per patient for evaluation. Median age was 45 years (interquartile range 35--54).

\textbf{Sex.} Sex was classified from ECG, ABP and PPG in VitalDB Open \cite{VitalDB} surgeries. Up to 30 windows of 60 s per surgery were encoded and averaged at the surgery level, and cross-validation folds were split at the patient level. The analysis set was 657 surgeries from 638 patients, 57.5\% male.

\subsection*{Supplementary Tables}

\begin{sidewaystable}
\caption{Overview of all datasets. ECG, electrocardiogram; ABP, arterial blood pressure; PPG, photoplethysmogram; CVP, central venous pressure; PAP, pulmonary artery pressure; ICP, intracranial pressure; \co, capnogram; AWP, airway pressure; RespImp, impedance respiration; RespFlow, respiratory flow; SNUH, Seoul National University Hospital; BIDMC, Beth Israel Deaconess Medical Center; MGH, Massachusetts General Hospital; OR, operating room; ICU, intensive care unit; \etco, end-tidal \co; MT, massive transfusion; DUA, data use agreement; MIMIC-III, Medical Information Mart for Intensive Care III; WDB, Waveform Database; KHDP, Korea Health Data Platform (research data platform operated by SNUH); IOH, intraoperative hypotension; PPV, pulse pressure variation; SVV, stroke volume variation; MAP, mean arterial pressure; LOS, length of stay; RR, respiratory rate.}\label{tab:datasets}
\scriptsize
\setlength{\tabcolsep}{3pt}
\begin{tabular}{@{}lllllp{4.2cm}p{2.2cm}p{4.0cm}@{}}
\toprule
Dataset & Institution & Country & Setting & N & Signals used & Access & Use \\
\midrule
SNUH OR Waveform Cohort & SNUH & Korea & OR & 75,567 records & ECG, ABP, PPG, CVP, PAP, \co, AWP, RespImp, RespFlow & Private & Pretraining \\
SNUH ICU Waveform Cohort & SNUH & Korea & ICU & 256,346 records & ECG, ABP, PPG, CVP, PAP, \co, AWP, RespImp, RespFlow & Private & Pretraining \\
K-MIMIC-MORTAL \cite{KMIMIC} & SNUH & Korea & ICU & 725 patients $\cdot$ 194,148 records & ECG, ABP, PPG, CVP, PAP, ICP, \co, AWP, RespImp, RespFlow & Restricted (KHDP) & Pretraining \\
VitalDB Open \cite{VitalDB} & SNUH & Korea & OR & 6,388 records $\cdot$ 6,090 patients & ECG, ABP, PPG, \co, AWP & Open (PhysioNet) & Downstream (IOH, \etco{} Derangement, PPV, SVV, MAP, Sex) \\
VitalDB Arrhythmia DB \cite{VitalDBArrhythmia} & SNUH & Korea & OR & 482 records & ECG, PPG & Open (PhysioNet) & Downstream (Arrhythmia) \\
SNUH MT Cohort & SNUH & Korea & OR & 17,986 records & ABP, PPG & Private & Downstream (Massive Transfusion) \\
SNUH Ventilation Cohort & SNUH & Korea & OR & 188 records & \co, AWP, RespFlow & Private & Downstream (Hypoxemia) \\
MIMIC-III WDB \cite{MIMICIIIWDB} & BIDMC & USA & ICU & matched 10,282 patients & ECG, ABP, PPG, CVP, PAP, ICP, RespImp & Open (PhysioNet) & Downstream (Cardiac Output, Intracranial Hypertension, Cardiac Arrest, ICU LOS, In-Unit Mortality, Mechanical Ventilation, RR) \\
MIMIC-III-Ext-PPG \cite{MIMICExtPPG} & BIDMC & USA & ICU & 6,189 patients & PPG & Credentialed (PhysioNet) & Downstream (Atrial Fibrillation) \\
CinC 2018 \cite{CinC2018} & MGH & USA & Sleep laboratory & 994 patients $\cdot$ 196,972 windows & Airflow, Chest, Abd & Open (PhysioNet) & Downstream (Sleep Apnea) \\
Aurora-BP \cite{AuroraBP} & Microsoft & USA & Wearable & 1,209 patients $\cdot$ 20,688 windows & ECG, PPG & DUA-restricted (Zenodo) & Downstream (Age) \\
\bottomrule
\end{tabular}
\end{sidewaystable}

CARMEN is reported at all three sizes (Small 52M, Base 174M, Large 410M); the three models were trained on the same data with the same training settings and the same number of tokens. Significance tests against comparison models and the figures quoted in the main text are based on CARMEN-Base.
\FloatBarrier

\begin{table}[!htbp]
\caption{\textbf{Clinical event detection.} For the 4-class Arrhythmia task the macro-average is reported. CARMEN, CArdio-Respiratory Multi-modal ENcoder; AUROC, area under the receiver operating characteristic curve; AUPRC, area under the precision--recall curve; CI, confidence interval. Bold, best value per task; underlined, second best.}\label{tab:s-detection}
\begin{tabular*}{\textwidth}{@{\extracolsep\fill}llcc}
\toprule
Task & Model & AUROC (95\% CI) & AUPRC (95\% CI) \\
\midrule
Arrhythmia & CARMEN-Large & \textbf{0.981} (0.971--0.989) & \underline{0.882} (0.831--0.919) \\
 & CARMEN-Base & \underline{0.978} (0.967--0.988) & \textbf{0.887} (0.835--0.924) \\
 & CARMEN-Small & 0.976 (0.965--0.986) & 0.863 (0.803--0.908) \\
 & MOMENT & 0.948 (0.932--0.964) & 0.764 (0.699--0.813) \\
 & NormWear & 0.929 (0.907--0.949) & 0.715 (0.640--0.778) \\
 & SPOTR & 0.875 (0.848--0.900) & 0.598 (0.541--0.662) \\
 & CARMEN-scratch & 0.943 (0.921--0.961) & 0.704 (0.642--0.784) \\
 & 1D-ResNet & 0.928 (0.904--0.947) & 0.752 (0.708--0.801) \\
\midrule
Hypoxemia & CARMEN-Large & 0.730 (0.676--0.784) & \textbf{0.072} (0.041--0.114) \\
 & CARMEN-Base & 0.774 (0.725--0.824) & \underline{0.068} (0.043--0.102) \\
 & CARMEN-Small & \textbf{0.806} (0.764--0.843) & 0.063 (0.038--0.099) \\
 & MOMENT & 0.746 (0.677--0.808) & 0.048 (0.029--0.079) \\
 & NormWear & 0.723 (0.655--0.789) & 0.040 (0.024--0.069) \\
 & SPOTR & 0.644 (0.588--0.704) & 0.026 (0.017--0.044) \\
 & CARMEN-scratch & \underline{0.797} (0.744--0.844) & 0.061 (0.039--0.086) \\
 & 1D-ResNet & 0.678 (0.614--0.745) & 0.031 (0.019--0.049) \\
\midrule
Atrial Fibrillation & CARMEN-Large & \underline{0.951} (0.939--0.962) & \underline{0.938} (0.920--0.952) \\
 & CARMEN-Base & \textbf{0.953} (0.942--0.962) & \textbf{0.941} (0.925--0.954) \\
 & CARMEN-Small & 0.949 (0.937--0.958) & 0.934 (0.917--0.948) \\
 & MOMENT & 0.915 (0.900--0.928) & 0.886 (0.860--0.908) \\
 & NormWear & 0.906 (0.890--0.920) & 0.868 (0.842--0.892) \\
 & SPOTR & 0.788 (0.769--0.805) & 0.682 (0.644--0.717) \\
 & CARMEN-scratch & 0.896 (0.880--0.912) & 0.849 (0.816--0.877) \\
 & 1D-ResNet & 0.908 (0.891--0.923) & 0.872 (0.844--0.896) \\
\midrule
Cardiac Output & CARMEN-Large & \underline{0.651} (0.620--0.681) & \textbf{0.317} (0.276--0.367) \\
 & CARMEN-Base & \textbf{0.656} (0.628--0.683) & \underline{0.315} (0.276--0.363) \\
 & CARMEN-Small & 0.647 (0.619--0.674) & 0.305 (0.266--0.350) \\
 & MOMENT & 0.641 (0.610--0.671) & 0.299 (0.261--0.350) \\
 & NormWear & 0.597 (0.567--0.630) & 0.262 (0.227--0.304) \\
 & SPOTR & 0.531 (0.502--0.560) & 0.229 (0.197--0.264) \\
 & CARMEN-scratch & 0.459 (0.424--0.500) & 0.191 (0.165--0.223) \\
 & 1D-ResNet & 0.585 (0.553--0.617) & 0.262 (0.224--0.306) \\
\midrule
Sleep Apnea & CARMEN-Large & \textbf{0.914} (0.909--0.918) & \textbf{0.784} (0.770--0.798) \\
 & CARMEN-Base & \underline{0.909} (0.903--0.914) & \underline{0.773} (0.758--0.788) \\
 & CARMEN-Small & 0.908 (0.902--0.912) & 0.766 (0.751--0.781) \\
 & MOMENT & 0.820 (0.812--0.829) & 0.575 (0.556--0.596) \\
 & NormWear & 0.826 (0.818--0.834) & 0.582 (0.561--0.604) \\
 & SPOTR & 0.848 (0.841--0.855) & 0.638 (0.618--0.658) \\
 & CARMEN-scratch & 0.904 (0.899--0.909) & 0.766 (0.751--0.780) \\
 & 1D-ResNet & 0.892 (0.886--0.897) & 0.731 (0.713--0.747) \\
\bottomrule
\end{tabular*}
\end{table}

\begin{table}[!htbp]
\caption{\textbf{Short-term event prediction.} Intraoperative Hypotension and Intracranial Hypertension predict a new event within 5 min after the end of observation, Massive Transfusion predicts massive transfusion beginning 5--10 min after the end of observation, and \etco{} Derangement predicts \etco{} derangement over the 1 min beginning 3 min after the end of observation. Bold, best value per task; underlined, second best. \etco, end-tidal \co; AUROC, area under the receiver operating characteristic curve; AUPRC, area under the precision--recall curve; CI, confidence interval.}\label{tab:s-prediction}
\begin{tabular*}{\textwidth}{@{\extracolsep\fill}llcc}
\toprule
Task & Model & AUROC (95\% CI) & AUPRC (95\% CI) \\
\midrule
Intraoperative Hypotension & CARMEN-Large & \textbf{0.895} (0.891--0.901) & \textbf{0.338} (0.323--0.354) \\
 & CARMEN-Base & \underline{0.895} (0.890--0.900) & \underline{0.336} (0.322--0.352) \\
 & CARMEN-Small & 0.888 (0.883--0.894) & 0.305 (0.292--0.319) \\
 & MOMENT & 0.664 (0.650--0.677) & 0.094 (0.086--0.104) \\
 & NormWear & 0.661 (0.647--0.675) & 0.097 (0.087--0.108) \\
 & SPOTR & 0.651 (0.639--0.664) & 0.089 (0.082--0.097) \\
 & CARMEN-scratch & 0.888 (0.883--0.893) & 0.321 (0.307--0.335) \\
 & 1D-ResNet & 0.814 (0.805--0.822) & 0.183 (0.169--0.197) \\
\midrule
Massive Transfusion & CARMEN-Large & \underline{0.869} (0.834--0.900) & \textbf{0.236} (0.171--0.319) \\
 & CARMEN-Base & \textbf{0.872} (0.836--0.903) & \underline{0.232} (0.172--0.314) \\
 & CARMEN-Small & 0.865 (0.832--0.896) & 0.201 (0.145--0.268) \\
 & MOMENT & 0.769 (0.728--0.808) & 0.091 (0.064--0.138) \\
 & NormWear & 0.839 (0.803--0.873) & 0.145 (0.104--0.201) \\
 & SPOTR & 0.796 (0.755--0.834) & 0.107 (0.078--0.153) \\
 & CARMEN-scratch & 0.666 (0.610--0.716) & 0.054 (0.038--0.080) \\
 & 1D-ResNet & 0.819 (0.776--0.856) & 0.139 (0.095--0.200) \\
\midrule
Intracranial Hypertension & CARMEN-Large & 0.856 (0.824--0.890) & 0.197 (0.149--0.253) \\
 & CARMEN-Base & \textbf{0.870} (0.838--0.900) & \underline{0.213} (0.162--0.284) \\
 & CARMEN-Small & 0.807 (0.768--0.851) & 0.163 (0.112--0.234) \\
 & MOMENT & 0.605 (0.549--0.662) & 0.040 (0.024--0.068) \\
 & NormWear & 0.557 (0.468--0.646) & 0.031 (0.019--0.054) \\
 & SPOTR & 0.551 (0.465--0.635) & 0.027 (0.018--0.039) \\
 & CARMEN-scratch & \underline{0.865} (0.820--0.908) & \textbf{0.242} (0.145--0.351) \\
 & 1D-ResNet & 0.798 (0.755--0.835) & 0.114 (0.072--0.172) \\
\midrule
\etco{} Derangement & CARMEN-Large & \underline{0.726} (0.706--0.744) & \underline{0.199} (0.177--0.223) \\
 & CARMEN-Base & 0.726 (0.706--0.746) & 0.198 (0.176--0.223) \\
 & CARMEN-Small & \textbf{0.737} (0.718--0.755) & \textbf{0.207} (0.183--0.234) \\
 & MOMENT & 0.669 (0.641--0.694) & 0.168 (0.147--0.196) \\
 & NormWear & 0.682 (0.657--0.705) & 0.180 (0.157--0.207) \\
 & SPOTR & 0.630 (0.606--0.653) & 0.151 (0.132--0.176) \\
 & CARMEN-scratch & 0.691 (0.669--0.714) & 0.152 (0.135--0.174) \\
 & 1D-ResNet & 0.651 (0.627--0.673) & 0.153 (0.131--0.181) \\
\bottomrule
\end{tabular*}
\end{table}

\begin{table}[!htbp]
\caption{\textbf{Admission outcome prediction.} The observation period is the first 3 h after ICU admission for Cardiac Arrest and ICU Length of Stay, 21--24 h after admission for In-Unit Mortality, and 1 h after the landmark for Mechanical Ventilation; window representations are averaged to form an admission-level representation (\hyperref[note:tasks]{Supplementary Note 1}). Because of the small samples, no supervised comparators were used, and models were compared with the same linear probe. Bold, best value per task; underlined, second best. ICU, intensive care unit; AUROC, area under the receiver operating characteristic curve; AUPRC, area under the precision--recall curve; CI, confidence interval.}\label{tab:s-outcome}
\begin{tabular*}{\textwidth}{@{\extracolsep\fill}llcc}
\toprule
Task & Model & AUROC (95\% CI) & AUPRC (95\% CI) \\
\midrule
Cardiac Arrest & CARMEN-Large & \underline{0.634} (0.568--0.696) & 0.471 (0.387--0.571) \\
 & CARMEN-Base & 0.634 (0.566--0.697) & \underline{0.493} (0.402--0.592) \\
 & CARMEN-Small & \textbf{0.636} (0.570--0.701) & \textbf{0.507} (0.417--0.608) \\
 & MOMENT & 0.510 (0.444--0.576) & 0.342 (0.278--0.426) \\
 & NormWear & 0.571 (0.502--0.640) & 0.452 (0.364--0.552) \\
 & SPOTR & 0.581 (0.511--0.646) & 0.441 (0.354--0.540) \\
\midrule
In-Unit Mortality & CARMEN-Large & 0.762 (0.730--0.791) & 0.443 (0.390--0.504) \\
 & CARMEN-Base & \underline{0.770} (0.739--0.797) & \underline{0.445} (0.392--0.507) \\
 & CARMEN-Small & \textbf{0.802} (0.774--0.828) & \textbf{0.490} (0.433--0.551) \\
 & MOMENT & 0.599 (0.562--0.634) & 0.265 (0.229--0.310) \\
 & NormWear & 0.721 (0.688--0.751) & 0.391 (0.341--0.451) \\
 & SPOTR & 0.693 (0.660--0.725) & 0.350 (0.305--0.407) \\
\midrule
Mechanical Ventilation & CARMEN-Large & 0.659 (0.619--0.698) & 0.206 (0.170--0.251) \\
 & CARMEN-Base & \underline{0.662} (0.623--0.698) & 0.202 (0.167--0.247) \\
 & CARMEN-Small & \textbf{0.678} (0.640--0.715) & \textbf{0.213} (0.177--0.258) \\
 & MOMENT & 0.587 (0.545--0.626) & 0.151 (0.127--0.185) \\
 & NormWear & 0.655 (0.616--0.691) & \underline{0.212} (0.171--0.262) \\
 & SPOTR & 0.650 (0.613--0.685) & 0.183 (0.150--0.226) \\
\midrule
ICU Length of Stay & CARMEN-Large & 0.675 (0.659--0.690) & 0.543 (0.519--0.567) \\
 & CARMEN-Base & \textbf{0.687} (0.672--0.702) & \underline{0.559} (0.536--0.584) \\
 & CARMEN-Small & \underline{0.684} (0.669--0.700) & \textbf{0.561} (0.537--0.585) \\
 & MOMENT & 0.585 (0.568--0.601) & 0.429 (0.408--0.452) \\
 & NormWear & 0.651 (0.635--0.667) & 0.522 (0.497--0.546) \\
 & SPOTR & 0.638 (0.622--0.653) & 0.498 (0.474--0.523) \\
\bottomrule
\end{tabular*}
\end{table}

\begin{table}[!htbp]
\caption{\textbf{Physiological estimation.} MAE and Pearson $r$ are reported. Within-subject $r$ is the correlation after removing the per-patient mean from both predictions and labels; it measures how well changes within a patient are tracked, independent of between-patient level differences, and can be higher or lower than the pooled $r$. MAE units are \% for PPV and SVV, mmHg for MAP and breaths/min for RR. Bold, best value per task; underlined, second best. PPV, pulse pressure variation; SVV, stroke volume variation; MAP, mean arterial pressure; RR, respiratory rate; MAE, mean absolute error; CI, confidence interval.}\label{tab:s-estimation}
\begin{tabular*}{\textwidth}{@{\extracolsep\fill}llccc}
\toprule
Task & Model & MAE & $r$ & Within-subject $r$ \\
\midrule
PPV & CARMEN-Large & \textbf{2.38} (2.16--2.61) & \textbf{0.625} & \textbf{0.468} \\
 & CARMEN-Base & \underline{2.52} (2.27--2.80) & 0.547 & 0.425 \\
 & CARMEN-Small & 2.54 (2.30--2.80) & \underline{0.548} & \underline{0.431} \\
 & MOMENT & 2.80 (2.52--3.11) & 0.436 & 0.375 \\
 & NormWear & 2.83 (2.56--3.14) & 0.384 & 0.346 \\
 & SPOTR & 3.20 (2.93--3.49) & 0.244 & 0.132 \\
 & CARMEN-scratch & 3.12 (2.86--3.39) & 0.368 & 0.340 \\
 & 1D-ResNet & 3.26 (3.00--3.50) & 0.393 & 0.344 \\
\midrule
SVV & CARMEN-Large & \textbf{2.70} (2.48--2.95) & \textbf{0.753} & \textbf{0.544} \\
 & CARMEN-Base & \underline{2.81} (2.55--3.11) & 0.717 & 0.501 \\
 & CARMEN-Small & 2.84 (2.59--3.12) & \underline{0.723} & \underline{0.503} \\
 & MOMENT & 3.22 (2.95--3.52) & 0.636 & 0.404 \\
 & NormWear & 3.46 (3.06--3.88) & 0.482 & 0.427 \\
 & SPOTR & 4.07 (3.66--4.55) & 0.148 & 0.058 \\
 & CARMEN-scratch & 3.94 (3.55--4.37) & 0.263 & 0.231 \\
 & 1D-ResNet & 4.15 (3.75--4.53) & 0.402 & 0.309 \\
\midrule
MAP & CARMEN-Large & \underline{9.67} (9.27--10.07) & \underline{0.572} & \underline{0.656} \\
 & CARMEN-Base & 9.82 (9.43--10.23) & 0.555 & 0.633 \\
 & CARMEN-Small & 9.96 (9.56--10.38) & 0.535 & 0.617 \\
 & MOMENT & 9.95 (9.56--10.35) & 0.538 & 0.620 \\
 & NormWear & 10.06 (9.64--10.47) & 0.518 & 0.637 \\
 & SPOTR & 10.97 (10.56--11.39) & 0.381 & 0.471 \\
 & CARMEN-scratch & 10.75 (10.34--11.18) & 0.435 & 0.504 \\
 & 1D-ResNet & \textbf{9.52} (9.05--10.00) & \textbf{0.573} & \textbf{0.674} \\
\midrule
RR & CARMEN-Large & \textbf{2.86} (2.65--3.06) & \textbf{0.784} & \textbf{0.528} \\
 & CARMEN-Base & \underline{3.21} (2.98--3.44) & \underline{0.754} & \underline{0.486} \\
 & CARMEN-Small & 3.41 (3.18--3.65) & 0.725 & 0.449 \\
 & MOMENT & 4.54 (4.24--4.80) & 0.526 & 0.238 \\
 & NormWear & 4.88 (4.59--5.16) & 0.446 & 0.219 \\
 & SPOTR & 5.42 (5.08--5.76) & 0.227 & 0.039 \\
 & CARMEN-scratch & 4.21 (3.93--4.50) & 0.561 & 0.303 \\
 & 1D-ResNet & 3.77 (3.50--4.04) & 0.650 & 0.384 \\
\bottomrule
\end{tabular*}
\end{table}

\begin{table}[!htbp]
\caption{\textbf{Phenotype.} In both tasks window-level predictions were averaged per patient for evaluation. For Age, NormWear takes only short fixed-length inputs, so the result of the same run as in Table~\ref{tab:fm} is reported. Age reports MAE (years) and Pearson $r$; Sex reports AUROC. MAE, mean absolute error; AUROC, area under the receiver operating characteristic curve; CI, confidence interval. Bold, best value per task; underlined, second best.}\label{tab:s-phenotype}
\begin{tabular*}{\textwidth}{@{\extracolsep\fill}llccc}
\toprule
Task & Model & MAE & $r$ & AUROC (95\% CI) \\
\midrule
Age & CARMEN-Large & \textbf{6.28} (6.03--6.54) & \textbf{0.743} & --- \\
 & CARMEN-Base & \underline{6.38} (6.11--6.64) & \underline{0.732} & --- \\
 & CARMEN-Small & 6.58 (6.31--6.85) & 0.718 & --- \\
 & MOMENT & 6.97 (6.68--7.25) & 0.658 & --- \\
 & NormWear & 7.21 (6.92--7.52) & 0.638 & --- \\
 & SPOTR & 8.26 (7.94--8.56) & 0.545 & --- \\
 & CARMEN-scratch & 7.66 (7.34--7.95) & 0.576 & --- \\
 & 1D-ResNet & 7.03 (6.74--7.32) & 0.639 & --- \\
\midrule
Sex & CARMEN-Large & --- & --- & \textbf{0.778} (0.742--0.814) \\
 & CARMEN-Base & --- & --- & \underline{0.776} (0.741--0.811) \\
 & CARMEN-Small & --- & --- & 0.766 (0.730--0.802) \\
 & MOMENT & --- & --- & 0.719 (0.680--0.757) \\
 & NormWear & --- & --- & 0.685 (0.646--0.725) \\
 & SPOTR & --- & --- & 0.673 (0.633--0.714) \\
 & CARMEN-scratch & --- & --- & 0.666 (0.623--0.710) \\
 & 1D-ResNet & --- & --- & 0.656 (0.614--0.699) \\
\bottomrule
\end{tabular*}
\end{table}

\begin{table}[!htbp]
\caption{Hyperparameters. MLP, multilayer perceptron; FFN, feed-forward network; LR, learning rate; AMP, automatic mixed precision; BF16, bfloat16; STFT, short-time Fourier transform; MPM, masked patch modeling; BNP, block next prediction; CMPM, cross-modal patch modeling; GLU, gated linear unit.}\label{tab:s-hyper}
\small
\begin{tabular*}{\textwidth}{@{\extracolsep\fill}lp{5.2cm}p{7.2cm}}
\toprule
Category & Hyperparameter & Value \\
\midrule
\textbf{Patch} & Patch size $P$ & 25 (0.25 s) \\
 & Stride & 25 (non-overlapping) \\
 & Residual MLP hidden dim & 50 ($2P$) \\
 & GLU FFN hidden dim & 2048 (Base; $\approx\tfrac{8}{3}d_{\mathrm{model}}$ rounded to a multiple of 8; Small 1,368, Large 2,736) \cite{Shazeer2020GLU} \\
\midrule
\textbf{Data pipeline} & Record window / crop range & 300 s / 10 s -- 300 s \\
 & Context length $L_{\mathrm{ctx}}$ (bin width) & Phase 1: 32,000 samples (1,280 tokens) / Phase 2: 48,000 samples (1,920 tokens) \\
 & Modality-balanced sampling & Temperature-based, $p_m \propto n_m^{0.5}$ ($s=0.5$), weight cap $5\times$ mean \\
 & Amplitude jitter $\eta_m$ & ECG 0.02 / other modalities 0.05 \\
\midrule
\textbf{Optimization} & Optimizer & AdamW ($\beta_1=0.9$, $\beta_2=0.95$, weight decay 0.1) \\
 & Learning rate (peak $\to$ final) & Phase 1: $2.0\times10^{-3}\to2.0\times10^{-4}$ / Phase 2: $3.5\times10^{-4}\to3.5\times10^{-5}$ \\
 & LR scheduler & Linear warmup $\to$ cosine decay to $0.1\times$ peak \\
 & Epochs $\times$ steps / warmup epochs & Phase 1: $12\times4{,}000$ / 1; Phase 2: $24\times500$, then $12\times3{,}000$ / 3 \\
 & Batch size (bins per GPU $\times$ GPUs) & Phase 1: $96\times7$ / Phase 2: $110\times7$ \\
 & Gradient clipping & 1.0 \\
 & Mixed precision & BF16 (AMP) \\
 & Seed & 42 \\
\midrule
\textbf{Objectives} & Mask ratio & 0.4 \\
 & Block-mask block size & 24--64 patches (6--16 s) \\
 & Next-prediction horizon $K$ & 24 patches (6 s) \\
 & $\alpha_{\mathrm{peak}}$ & 1.0 \\
 & STFT window set $\mathcal{N}$ / $\lambda_{\mathrm{spec}}$ & $\{8,16\}$ / 0.2 \\
 & $\lambda_{\mathrm{MPM}}$ / $\lambda_{\mathrm{BNP}}$ / $\lambda_{\mathrm{CMPM}}$ / $\lambda_{W}$ & 1.0 / 0.8 / 0.3 (Phase 2) / 0.005 (Phase 2) \\
 & Variate-level mask probability (Phase 2) & 0.3 \\
\midrule
\textbf{Linear probe} & Head & LayerNorm $\to$ Dropout(0.1) $\to$ Linear \\
 & Optimizer / LR / batch size & Adam / $1\times10^{-3}$ / 512 \\
 & Loss & Cross-entropy (classification) / MSE on standardized labels (regression) \\
 & Max epochs / early stopping & 2,000 ($\leq$ 300k steps) / patience 200, best validation epoch restored \\
 & Data split & Patient-level 5-fold; fold $k$ test, fold $(k+1)\bmod5$ validation \\
\bottomrule
\end{tabular*}
\end{table}

\newpage
\bibliographystyle{unsrtnat-nonote}   
\bibliography{library}
\end{document}